\documentclass[10pt,twocolumn,letterpaper]{article}

\usepackage{iccv}              

\input{mlpreamble}
\usepackage[accsupp]{axessibility}  %
\usepackage{xcolor}
\usepackage{color, colortbl}
\usepackage{amsmath}
\usepackage{mathtools}
\usepackage{empheq}
\usepackage{siunitx}
\usepackage{wrapfig}
\usepackage{cancel}
\usepackage[whole]{bxcjkjatype}
\usepackage{amssymb}
\usepackage{multirow}
\definecolor{Gray}{gray}{0.85}
\definecolor{LightGray}{gray}{0.95}
\definecolor{LightRed}{rgb}{0.98,0.859,0.85}
\definecolor{LightYellow}{rgb}{0.99,0.95,0.81}
\definecolor{LightOrange}{rgb}{0.99,0.92,0.816}
\definecolor{DarkRed}{rgb}{0.8,0,0}
\definecolor{LightBlue}{rgb}{0,0.44,0.753}

\def\onedot{.~}
 
\def\ie{\emph{i.e.},~}

\def\iow{i.o.w.,~}
\def\etal{\emph{et al}\onedot}

\newcommand{\doubleR}{\mathbb{R}}
\newcommand{\doubleRp}{\mathbb{R}_+}

\newcommand{\diag}[1]{\ensuremath{\mathrm{diag}\left({#1}\right)}}
\newcommand{\checkbox}[1]{%
  \ifnum#1=1
    \makebox[0pt][l]{\raisebox{0.15ex}{\hspace{0.1em}$\checkmark$}}%
  \fi
  $\square$
}

\newcommand{\ithEq}[1]{${#1}^\mathrm{th}$}

\newcommand{\syneos}{EOS~600D\xspace}
\newcommand{\syniphone}{iPhone12ProMax\xspace}
\newcommand{\synpentax}{Pentax~K5\xspace}
\newcommand{\synolympus}{Olympus~EPL2\xspace}
\newcommand{\synsumsung}{Sumsung~GalaxyS20\xspace}

\newcommand{\sony}{Sony~$\alpha$1\xspace}
\newcommand{\eos}{EOS~RP\xspace}
\newcommand{\iphonexv}{iPhone~15ProMax\xspace}
\newcommand{\djipocket}{DJI~Pocket3\xspace}
\newcommand{\cc}{\mbox{CC}\xspace}
\newcommand{\exif}{\mbox{Exif+CC}\xspace}

\definecolor{softred}{RGB}{255, 200, 200}
\definecolor{softyellow}{RGB}{255, 240, 180}

\newcommand{\bestscore}[2]{%
    \makebox[#1]{\underline{#2}}%
}
\newcommand{\secondbest}[2]{%
    \makebox[#1]{\underline{#2}}%
}

\newcommand{\raisedtarget}[1]{%
  \raisebox{\fontcharht\font}[0pt][0pt]{\hypertarget{#1}{}}%
}

\definecolor{iccvblue}{rgb}{0.21,0.49,0.74}
\usepackage[pagebackref,breaklinks,colorlinks,allcolors=iccvblue]{hyperref}

\def\paperID{5710} 
\def\confName{ICCV}
\def\confYear{2025}

\title{Spectral Sensitivity Estimation with an Uncalibrated Diffraction Grating}
\author{
Lilika~Makabe$^{1}$ \quad
Hiroaki~Santo$^{1}$ \quad
Fumio~Okura$^{1}$ \quad
Michael S. Brown$^{2}$\quad 
Yasuyuki~Matsushita$^{1,3}$
\\
$^{1}$The University of Osaka \quad
$^{2}$York University \quad
$^{3}$Microsoft Research Asia -- Tokyo\\
{\tt\small \{makabe.lilika,santo.hiroaki,okura,yasumat\}@ist.osaka-u.ac.jp} \quad
{\tt\small mbrown@yorku.ca}
}

\newcommand{\vsp}{-4mm}

\begin{document}
\maketitle
\begin{abstract}
This paper introduces a practical and accurate calibration method for camera spectral sensitivity using a diffraction grating. Accurate calibration of camera spectral sensitivity is crucial for various computer vision tasks, including color correction, illumination estimation, and material analysis. Unlike existing approaches that require specialized narrow-band filters or reference targets with known spectral reflectances, our method only requires an uncalibrated diffraction grating sheet, readily available off-the-shelf. By capturing images of the direct illumination and its diffracted pattern through the grating sheet, our method estimates both the camera spectral sensitivity and the diffraction grating parameters in a closed-form manner. Experiments on synthetic and real-world data demonstrate that our method outperforms conventional reference target-based methods, underscoring its effectiveness and practicality.
\end{abstract}
\section{Introduction and Preliminaries}
\label{sec:intro}

Estimating a camera's spectral sensitivity---its response to each wavelength of incident light---is essential for tasks such as color measurement, illumination and reflectance estimation, and material recognition. Spectral sensitivity is governed by the camera's optical system, including the sensor's color filter array and lens.

Conventional methods~\cite{jimaging10070169, darrodi2015reference, berra2015estimation, macdonald2015determining} for calibrating spectral sensitivity rely on specialized equipment to emit light with known spectra toward the camera and record the corresponding observed intensities. While accurate, these approaches are costly due to the need for precisely designed optical systems and extensive measurement time.
Rather than directly observing light sources with known spectra, prior works~\cite{jiang2013space,rump2011practical} have proposed using reference targets with known spectral reflectances under known spectral illuminations, such as sunlight or standard illuminants like D65. While these methods simplify the calibration process, a key challenge is the inherently low-frequency nature of spectral reflectances observed on natural objects~\cite{Parkkinen1989Characteristic} and high correlation between the spectral reflectances of individual patches~\cite{dicarlo2004emissive}, which makes it difficult to accurately determine sensitivity at individual wavelengths.

\begin{figure}
    \centering
    \includegraphics[width=\linewidth]{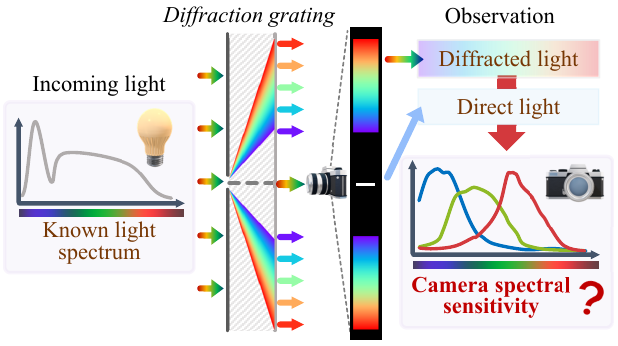}
    \caption{
 {Overview of the proposed method.} By capturing images of light with a known spectrum through an uncalibrated diffraction grating sheet, we observe both the direct and diffracted light. This practical and easy-to-use setup enables accurate estimation of camera spectral sensitivity.
 \vspace{-1em}
    }
    \label{fig:teaser}
\end{figure}

To enhance calibration accuracy, the use of diffraction grating sheets~\cite{karge2018using, toivonen2020practical} has been explored in the past. Diffraction occurs when light waves bend and spread as they pass through narrow apertures, with the degree of bending dependent on wavelength. This effect can be observed using an off-the-shelf diffraction grating sheet, available for less than $5$ USD. Each diffracted light beam produces distinct spectral peaks, enabling precise estimation of spectral sensitivities at individual wavelengths.
A key challenge in using a diffraction grating sheet is calibrating its diffraction parameter, the grating efficiency that describes the wavelength-dependent attenuation. To calibrate the efficiency, current methods require measurements of a reference target with known reflectances along with the diffracted light beam observation. This involves capturing multiple calibration scenes and complicates the process.

For an accurate and easy-to-use calibration method, we propose using an {\em uncalibrated} diffraction grating sheet without needing additional captures of reference targets. As shown in \fref{fig:teaser}, the proposed method 
observes light with known spectra through the diffraction grating sheet, recording both the direct light and the diffraction pattern.

The proposed method simultaneously estimates the camera spectral sensitivity and grating efficiency. Although the original problem is a bilinear problem where the unknowns appear in their product form, we show that it can be turned into a tractable linear problem by introducing a basis representation for both camera spectral sensitivity and grating efficiency. We further introduce a closed-form solution method to the problem, given the pixel-to-wavelength mapping, \ie the mapping between the pixel locations in the captured image and the wavelengths of the diffracted light.

We conduct both synthetic and real-world experiments to demonstrate the effectiveness of the proposed method. The results show that our method outperforms existing leading methods using calibrated reflectance targets~\cite{jiang2013space}.

\begin{figure}
 \vspace{-1em}
    \centering
    \includegraphics[width=\linewidth]{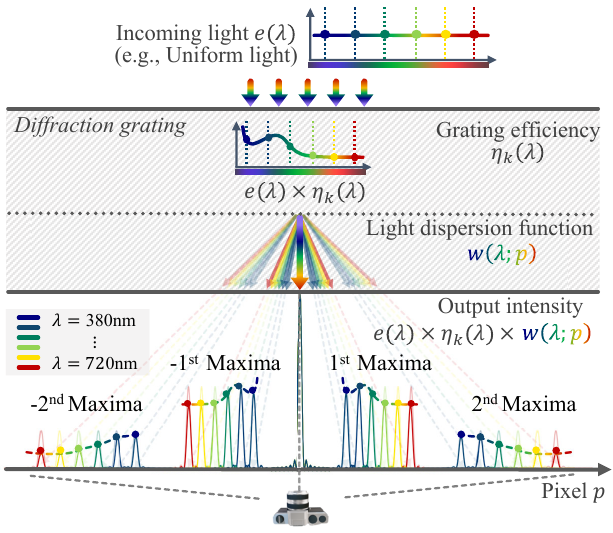}
    \\
    \vspace{-0.8em}
    \caption{
    Illustration of a diffraction grating.
    When light of a certain wavelength passes through a diffraction grating, its output intensity is spatially distributed in a characteristic pattern with multiple sharp peaks, known as intensity principal maxima.
    The pixel positions of the intensity principal maxima shift outward from the center for longer wavelengths. When a broadband light source passes, light beams of different wavelengths appear at distinct shifted positions.
    In practice, we also need to consider non-uniform transmittance, known as grating efficiency.
    As a result, the observed intensity at a given pixel position is the product of the incoming light spectrum, the light dispersion function, and the grating efficiency. See ~\fref{fig:setup-and-observation} for examples of the observations.
    \vspace{-0.6em} 
    }
    \label{fig:diffraction_ex}
\end{figure}


\vspace{\vsp}
\paragraph{Contributions} Our chief contributions are as follows:
\begin{itemize}
\item We propose a camera spectral sensitivity estimation method that uses an uncalibrated diffraction grating sheet, without the need for capturing reference targets.
\item The method uses an inexpensive diffraction grating sheet,
making it practical, cost-effective, and easy to use.
\item Accuracy of the proposed method significantly outperforms those that rely on reference targets, with a simpler setup and fewer captures.
\end{itemize}

\section*{Preliminaries}
\label{sec:background}
In this section, we provide the foundational background on spectral image formation. We introduce the general spectral image formation model, followed by an overview of how diffraction gratings separate light into its constituent wavelengths. Finally, we present the image formation model for observations captured through a diffraction grating.

\vspace{\vsp}
\paragraph{Spectral image formation}
Given camera spectral sensitivity $s(\lambda_i)$ at discretely sampled wavelengths $\{\lambda_i\}$ and spectrum of the incoming light~$e(\lambda_i)$ to the camera, the intensity observation $m\in\doubleRp$ can be expressed as:
\begin{equation}
    \label{eq:imageformation}
    m \propto \sum_{i} e(\lambda_i) s(\lambda_i).
\end{equation}

\vspace{\vsp}
\paragraph{Diffraction grating}

\begin{figure}
    \centering
    \includegraphics[width=\linewidth]{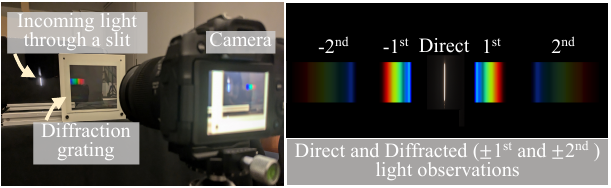}
    \caption{
    Capturing setup and the observed image of direct and diffracted lights.
    As the diffracted light exhibits multiple principal maxima, multiple diffraction patterns can be observed. The $k$-th diffraction image, counted from the center ($k = \pm1, \pm2, \dots$), is referred to as the \textit{$k$-th order diffraction image}. The image intensity is adjusted for visualization. 
    } 
    \label{fig:setup-and-observation}
\end{figure}

A diffraction grating is an optical device with multiple periodic grooves that separates incoming light into its constituent wavelengths, directing each along a distinct path. Due to interference from the multi-groove structure, each wavelength produces multiple principal intensity maxima, corresponding to different diffraction orders $k = \pm1, \pm2, \dots$, as illustrated in \fref{fig:diffraction_ex}. This results in multiple dispersed observations symmetrically aligned with the direction of incoming light, with the central observation representing the mixture of all constituent wavelengths—namely, the direct light observation.

In general, diffraction observation $m_c^{(k)}(p)$
around the \ithEq{k} intensity principal maxima (\iow \ithEq{k}~order diffraction observation)
at pixel location~$p$ in channel $c \in \{R,G,B\}$ can be written as
\begin{equation}
    \label{eq:1st-image-formation}
    m_c^{(k)}(p) \propto \sum_i w(\lambda_i; p) e(\lambda_i
)\eta_k(\lambda_i) s_c(\lambda_i),
\end{equation}
where $s_c$ is the camera spectral sensitivity for channel $c$, and $\eta_k(\lambda)$ represents the grating efficiency for each wavelength $\lambda$ at $k^{\mathrm{th}}$ order diffraction. The term $w(\lambda; p)$ is the weight induced by the light dispersion function, which disperses the light into each wavelength $\lambda$ that is observed at a corresponding peak pixel position $p$ depending on the wavelength. We refer to $w(\lambda;p)$ as a weight function.

The weight function $w(\lambda;p)$ approximates a delta function with a sufficient number of slits in the grating, effectively acting as a one-to-one mapping between wavelength $\lambda$ and pixel position $p$ in the diffraction image~\cite{karge2018using, toivonen2020practical}. Ideally, the grating efficiency $\eta(\lambda)$ should be all $1$. However, in practice, the grating efficiency is known to vary depending on the frequency and the material of the grating~\cite{palmer2005diffraction}.

\section{Related Work}
\label{sec:relatedwork}
In this section, we provide an overview of existing methods for camera spectral sensitivity calibration. A straightforward approach to recovering spectral sensitivity is to measure the observed intensities $m$ in the image while varying the incoming light with known spectra $e(\lambda_i)$. To achieve such controlled illuminations, specialized light sources~\cite{hubel1994comparison} and light sources with narrow-band filters~\cite{jimaging10070169, darrodi2015reference, berra2015estimation, macdonald2015determining} have been used. However, the equipment is expensive, limiting their practical applications in real-world settings.

To eliminate the need for specialized light sources,
methods using reference targets, such as color checker charts~\cite{alsam2002recovering, quan2003comparative, ebner2007estimating, rump2011practical, prasad2013quick, finlayson2016rank, ji2021compressive, tominaga2021measurement, kaya2019towards}, have been proposed. These methods recover camera spectral sensitivity by observing reference targets composed of multiple surface patches with known spectral reflectances under calibrated spectral illumination that uniformly illuminates the entire target.

To reduce the cost of spectral illumination calibration, several methods~\cite{dyas2000robust, kawakami2013camera} assume specific illumination types, such as sunlight~\cite{kawakami2013camera} or low-frequency spectra~\cite{huynh2014recovery, zhang2017camera}, and simultaneously estimate camera spectral sensitivity and illumination spectra. 
While these methods can measure reflected light with different spectra from multiple patches in a single image, they rely on reference targets with known spectral reflectance. Unfortunately, off-the-shelf color checker charts do not provide such data, so one must either rely on externally measured data or perform their own calibration~\cite{zhu2020spectral}.
In addition, due to the low-frequency spectral reflectance of natural objects, the spectral resolution of these methods is limited. To address this challenge, methods using emissive charts~\cite{dicarlo2004emissive} have been proposed; however, reference targets employing fluorescent materials are rare and have limited practical applicability.

Instead of observing reflectances, methods using a diffraction grating, which makes light beams with distinct spectral peaks at different wavelengths, have been proposed~\cite{karge2018using,toivonen2020practical}.
Karge~\etal~\cite{karge2018using} uses a fluorescent lamp, which has a spiky spectrum, to calibrate the pixel-to-wavelength mapping. They then replace the fluorescent lamp with a halogen lamp, which has a flatter spectrum, to observe the diffraction pattern. They additionally need to capture reference targets with known spectral reflectances under controlled spectral illumination to account for the attenuation due to grating efficiency.
Toivonen~\etal~\cite{toivonen2020practical} use multiple light sources with different spectra, generated by a halogen lamp equipped with calibrated color filters, to capture the multiple diffraction patterns. In addition, they also record images of a transmissive color chart, comprising multiple patches of calibrated color filters, through different light sources.
By recording the diffraction patterns of different light sources and images of the transmissive color chart as input, they achieve accurate calibration of the spectral sensitivity. 
These methods use inexpensive setups compared to the specialized equipment and achieve more accurate calibration than those relying solely on a color checker chart. However, these approaches require multiple setups and complex capture procedures, which remain a significant practical challenge. 

Unlike the above approaches, Solomatov and Akkaynak~\cite{solomatov2023spectral} use metadata recorded in the Exif, specifically the color matrices that transforms sensor responses to the CIE XYZ color space, to train a neural network for estimating spectral sensitivity. This approach does not require any additional captures for calibration; however, it cannot account for external factors, such as lens and camera filters. Moreover, their estimation exhibits white-balance ambiguity, and recovering the full spectral sensitivity requires white-balance calibration using a color checker chart.

We use an uncalibrated diffraction grating sheet, where the grating efficiency is unknown. Unlike previous diffraction grating-based methods, our approach only requires capturing a light source through the diffraction grating sheet, providing both the direct light and the diffraction pattern, enabling practical and accurate estimation. 
Notably, our minimal setting uses one light and two images under the light, and does not require any scene interaction after setup, whereas existing methods require at least two scene changes and three lights with known spectra~\cite{karge2018using}.
\section{Proposed Method}
\label{sec:problem_statement}
\begin{figure*}
\vspace{-4mm}
    \centering
    \includegraphics[width=\linewidth]{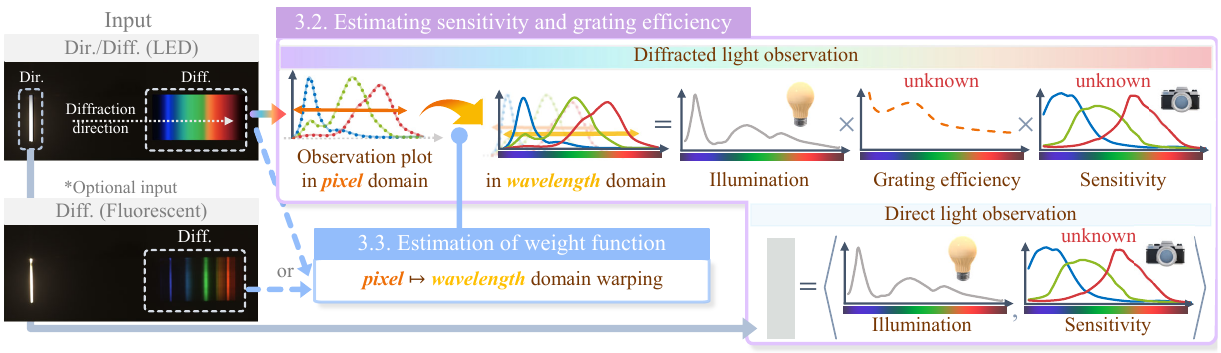}
    \caption{
Illustration of the image formation models and overview of our estimation.
The observation should vary only along the diffraction direction. To obtain the mean observation, we average along the vertical direction in the figure, resulting in the direct light observation in $\doubleR^{3}$ and the diffracted light observation in $\doubleR^{n \times 3}$ in RGB channels, respectively. The sampled observation plot in the bottom figure shows a per-channel vector plot along the diffraction direction for each channel. This plot should have the same outline as the wavelength-wise multiplication of the grating efficiency, camera spectral sensitivity, and the incoming light spectrum, leaving the unknown mapping along the horizontal axis, referred to as the pixel-to-wavelength mapping. Using both the direct and diffracted light observation, given the pixel-to-wavelength mapping, we estimate the camera spectral sensitivity and the grating efficiency as described in Sec~\ref{sec:camspecs-est}. Section~\ref{sec:pix2wave} covers the estimation of the pixel-to-wavelength mapping.
}
    \label{fig:method}
\end{figure*}

In this section, we first outline the problem setting for determining the camera spectral sensitivity using a diffraction grating. We then present the proposed method.

\vspace{\vsp}
\paragraph{Problem setting}

Our image formation models of the direct light observation $m_\mathrm{dir}$ and the first-order diffracted light observation $m_\mathrm{dif}$ are written as :
\begin{equation}
    \label{eq:image-formation}
    \begin{aligned}
        m_\mathrm{dir} &\propto E \sum_i e(\lambda_i) s(\lambda_i),\\
        m_\mathrm{dif}(p) &\propto E \sum_i w(\lambda_i; p) e(\lambda_i) \eta(\lambda_i) s(\lambda_i) ,
    \end{aligned}
\end{equation}
where $E$ is an unknown scale of the incoming light spectrum and $e(\lambda)$ is a normalized spectrum of the incoming light. We omit the pixel position dependency for the direct light observation $m_\mathrm{dir}$ as it remains constant along the diffraction (horizontal) direction, unlike the diffracted light. We do not use $k\ge2$ diffracted light observations in our method because, in practice, the intensity drastically decreases for higher-order diffraction. 

Given a known incoming light spectrum $e(\lambda)$ and by observing both the direct and first-order diffracted lights, $m_\mathrm{dir}$ and $m_\mathrm{dif}(p)$, our goal is to estimate the camera's spectral sensitivity $s(\lambda)$ using an uncalibrated diffraction grating, where both the weight function $w(\lambda;p)$ and the grating efficiency $\eta(\lambda)$ are unknown.

In the following, we first describe the method to jointly estimate the camera spectral sensitivity $s(\lambda)$ and the grating efficiency $\eta(\lambda)$, assuming that the weight function $w(\lambda;p)$ is known beforehand. We then present the method to determine the weight function $w(\lambda;p)$.

\subsection{Estimating camera spectral sensitivity $s$ and grating efficiency $\eta$} 
\label{sec:camspecs-est}
We now estimate the camera spectral sensitivity $s(\lambda)$ and the grating efficiency $\eta(\lambda)$, given the weight function $w(\lambda; p)$. We show, for the first time, that the problem has a closed-form solution when the unknowns are expressed as linear combinations of their bases. To this end, we first explain the constraints derived from the image formation models of the direct and diffracted light observations, and then present the solution method.

Let us assume camera spectral sensitivity $\V{s}$ and the inverse of grating efficiency $\boldsymbol{\eta}^{-1}$ at discretely sampled wavelengths $\lambda_1, \ldots, \lambda_f$ can be represented by linear combinations of bases whose numbers of bases are $b_s$ and $b_\eta$, respectively. Namely, 
\begin{equation}
    \begin{aligned}
        \label{eq:basis-rep}
        \V{s} &\triangleq
        \begin{bmatrix}
            s(\lambda_1), & \ldots, & s(\lambda_f)
        \end{bmatrix}\transp 
        = \V{B}_s\V{c}_s  \in \doubleRp^{f}, \\
        \boldsymbol{\eta}^{-1} 
        &\triangleq 
        \begin{bmatrix}
            \frac{1}{\eta(\lambda_1)}, & \ldots,& \frac{1}{\eta(\lambda_f)}
        \end{bmatrix}\transp 
        = \V{B}_\eta\V{c}_\eta 
        \in \doubleRp^{f},
    \end{aligned}
\end{equation}
where \hbox{$\V{B}_\mathrm{*}\in\doubleR^{f\times b_\mathrm{*}}$} is a basis matrix, and $\V{c}_*\in\doubleR^{b_*}$ is a coefficient vector. 
We obtained the basis matrix $\V{B}_s$ of camera spectral sensitivity $\V{s}$ at a channel by singular value decomposition~(SVD) on the same set of data used in~\cite{solomatov2023spectral}.
For the inverse grating efficiency function $\boldsymbol{\eta}^{-1}$, we use a Fourier basis representation. Further details can be found in the supplementary material.

\vspace{\vsp}
\paragraph{Linear constraint from direct light observation}
By introducing the basis representation, the aforementioned image formation model of direct light observations in \eref{eq:image-formation} can be rewritten in a vector form as 
\begin{equation}
    \label{eq:dir-image-matrix}
    \begin{aligned}
        m_\mathrm{dir} &\propto E\V{e}\transp \V{s} = E\V{e}\transp \V{B}_s\V{c}_s,
    \end{aligned}
\end{equation}
where $\V{e}\in\doubleRp^{f}$ is the normalized incoming light spectrum.
The light scale $E$ can be set to $1$, and the proportionality can be turned into equality without loss of generality, as the global scale can be accounted for in camera spectral sensitivity $\V{s}$. As a result, we have 
\begin{equation}
 m_\mathrm{dir}=\underbrace{\V{e}\transp \V{B}_s}_{\V{A}_\mathrm{dir}}\V{c}_s,
\end{equation}
where \hbox{$\V{A}_\mathrm{dir} \in \doubleR^{1\times b_s}$} is a known row vector. Similar equations can be derived for all RGB channels to form a vector of direct light observations \hbox{$\V{m}_\mathrm{dir}=[m_{\mathrm{dir}(R)}, m_{\mathrm{dir}(G)}, m_{\mathrm{dir}(B)}]\transp$} on the left-hand side of the equation, resulting in $3$ linear equations with unknown vector $[\V{c}_{s(R)}\transp, \V{c}_{s(G)}\transp, \V{c}_{s(B)}\transp]\transp$, for which $\V{A}_\mathrm{dir}$ becomes a~$3\times 3b_s$ matrix.

\vspace{\vsp}
\paragraph{Linear constraint from diffracted light observation}
Like the direct light observation case, the diffracted light's image formation model in \eref{eq:image-formation} can be rewritten in a matrix form as
\begin{equation}
    \label{eq:dif-image-matrix}
    \begin{aligned}
        \V{m}_\mathrm{dif} &= 
        \begin{bmatrix}
            m_\mathrm{dif}(p_1), & \ldots, & m_\mathrm{dif}(p_n)
        \end{bmatrix}\transp \\
        &= 
        \V{W}
        \diag{\V{e}}\diag{\boldsymbol{\eta}}
        \V{s},
    \end{aligned}
\end{equation}
where 
$\V{m}_\mathrm{dif}\in\doubleRp^n$ is a stack of the diffracted light observations at pixels $p_1,\ldots,p_n$. $\V{W}\in\doubleRp^{n\times f}$
denotes the known weight function for all $n$ pixels, written in a matrix form as 
\begin{align}
        \V{W} &= 
        \begin{bmatrix}
            w(\lambda_1, p_1) & \dots & w(\lambda_f, p_1)\\
            \vdots \\
            w(\lambda_1, p_n) & \dots &  w(\lambda_f, p_n)\\
        \end{bmatrix} \in \doubleRp^{n\times f}, 
\end{align}
which we refer to as the weight matrix.

The above image formation model of diffracted light can be equivalently turned into 
\begin{equation}
    \begin{aligned}
    \V{s} &=  \diag{\boldsymbol{\eta}^{-1}}\underbrace{\diag{\V{e}^{-1}} \V{W}^\dagger\V{m}_\mathrm{dif} }_{\V{a}} 
    \label{eq:dif-image-const-nonlinear}
    = \diag{\V{a}}\boldsymbol{\eta}^{-1},
    \end{aligned}
\end{equation}
where the inverse of a vector denotes an element-wise inverse operator, $\V{W}^\dagger$ is a left pseudo-inverse of $\V{W}$, and $\V{a}\in\doubleR^f$ is a known channel-and-light-dependent vector. Since we can confidently assume $n > f$ in our setting, the weight matrix $\mathbf{W}$ should have full column rank when the diffracted observation spans the entire visible wavelength domain, and thus it has a left pseudo-inverse.

From \eref{eq:basis-rep}, \eref{eq:dif-image-const-nonlinear} can be written as
 \begin{equation}
    \V{B}_s\V{c}_s = \diag{\V{a}}\V{B}_\eta\V{c}_\eta. 
 \end{equation}
Putting together the unknowns $\V{c}_\eta$ and $\V{c}_s$, we have the following homogeneous system of equations:
\begin{equation}
    \underbrace{\begin{bmatrix}
        \diag{\V{a}}\V{B}_\eta &-\V{B}_s  
        \end{bmatrix}}_{\V{A}_\mathrm{dif}}
    \begin{bmatrix}
    \V{c}_\eta \\
    \V{c}_s 
    \end{bmatrix}
    = \V{0},
\end{equation}
where \mbox{$\V{A}_\mathrm{dif}\in \doubleR^{f\times (b_\eta+ b_s)}$} is a known matrix.
For the case of RGB images, stacking similar equations from all RGB channels, we obtain $3f$ linear equations from the diffracted light observations for unknown vector \mbox{$[\V{c}_\eta\transp,  \V{c}_{s(R)}\transp,  \V{c}_{s(G)}\transp,  \V{c}_{s(B)}\transp]\transp\in\doubleR^{b_\eta + 3b_s}$}, and \mbox{$\V{A}_\mathrm{dif}$} becomes a~$3f \times (b_\eta+3b_s)$ matrix.

\vspace{\vsp}
\paragraph{Solution method}\label{sec:solution}

We can estimate the camera spectral sensitivity $\V{s}$ and grating efficiency $\boldsymbol{\eta}$ by solving the following optimization problem using RGB channels:
\begin{equation}
    \begin{gathered}
    \V{x}^* = \argmin_{\V{x}} \left\| \V{A}_\mathrm{dif} 
\V{x} \right\|_2^2~~
\mathrm{s.t.}~~ 
    \begin{bmatrix}
        \V{0} & \V{A}_\mathrm{dir}
    \end{bmatrix}    
\V{x} = 
\V{m}_\mathrm{dir},
    \end{gathered}
\end{equation}
where 
$\V{x} = \left[{\V{c}_\eta}\transp,  \V{c}_{s(R)}\transp,  \V{c}_{s(G)}\transp, \V{c}_{s(B)}\transp \right] \transp\in\doubleR^{b_\eta + 3b_s}$
is the vector of unknowns.
The above optimization problem has a closed-form solution by introducing a Lagrange multiplier when the number bases $b_*$ and the wavelength resolution $f$ satisfy $3f \ge b_\eta + 3b_s$, and $\V{A}_\mathrm{dif}$ and $\V{A}_\mathrm{dir}$ are full-rank.  
The details of the solution method are provided in the supplementary material. 

\subsection{Estimation of weight function $w$}
\label{sec:pix2wave}
\begin{figure}
    \centering
    \includegraphics[width=\linewidth]{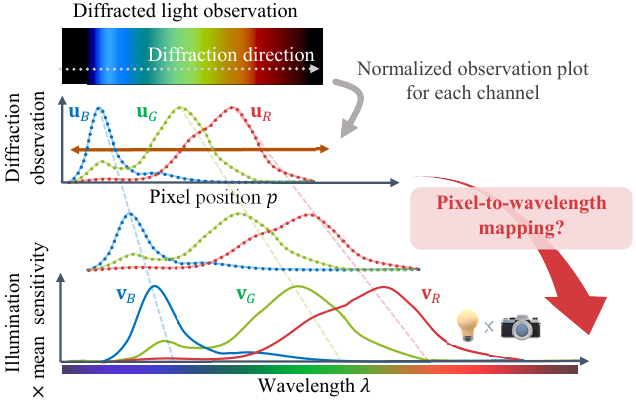}
    \caption{
       Illustration of our pixel-to-wavelength mapping estimation. 
       Given the normalized diffraction observations $\V{u}_c$ and normalized illumination-scaled mean camera spectral sensitivity $\V{v}_c$ for all $c\in\{R,G,B\}$ channels, we estimate the optimal pixel-to-wavelength mapping (or horizontal warping, as shown in the figure), which minimizes the distance between the two plots.
    }
    \vspace{-1.em}
    \label{fig:pix2wave_est}
\end{figure}

As described in \sref{sec:background}, the weight function $w(\lambda ; p)$ has a peaky shape using diffraction gratings; therefore, we assume there is a one-to-one mapping between the pixel location $p$ in the diffraction image and the wavelength $\lambda$ as in previous methods~\cite{karge2018using, toivonen2020practical}.
To estimate the weight function $w(\lambda;p)$, it is necessary to establish the pixel-to-wavelength mapping, namely, relating the pixel locations of the diffraction image to the wavelength as illustrated in \fref{fig:pix2wave_est}.

We propose to use a quadratic function to represent the one-to-one pixel-to-wavelength mapping $g: p \rightarrow \lambda$ as
\begin{equation}
    \label{eq:gaussian-approx}
    \lambda = g(p) = ap^2 + bp + c,
\end{equation}
where $a,b,c\in\doubleR$ are the parameters. This quadratic representation allows us stably determining the mapping with retaining accuracy.
For further discussion about this parameterization, please refer to our supplementary.

Once the pixel-to-wavelength mapping $g(p)$ is determined, the elements of the weight matrix $\V{W}$ is simply obtained by linear interpolation as
\begin{equation}
    w(\lambda, p) = \left\{
        \begin{aligned}
            & \frac{\lambda_{i+1} - g(p)}{\lambda_{i+1} - \lambda_{i}} && \text{if } \lambda = \lambda_i\\
            & \frac{g(p) - \lambda_{i}}{\lambda_{i+1} - \lambda_{i}} && \text{if } \lambda = \lambda_{i+1}\\
            &0  && \text{otherwise}
        \end{aligned}
    \right. ,
\end{equation} 
where 
$\lambda_i$ is the nearest-neighbor wavelength to $g(p)$ that satisfies $\lambda_i \le g(p) \le \lambda_{i+1}$. Therefore, the problem of estimating the weight function boils down to the estimation of pixel-to-wavelength mapping.
The challenge, though, is that we only have access to the diffraction image and illumination spectra, but not to camera spectral sensitivity and grating efficiency. Now we describe two settings for determining the pixel-to-wavelength mapping; one using fluorescent and LED lights (denoted as Fluorescent+LED), and another using only LED light (denoted as LED only).

\begin{figure}
    \centering
    \includegraphics[width=\linewidth]{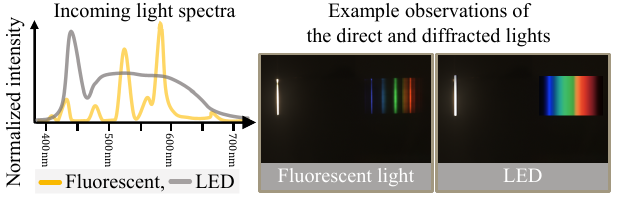}
    \caption{
    The illuminant spectrum of the light source used in our real-world experiment and an example of the corresponding direct and diffracted light observations. The observations shown here are adjusted for visualization.
    }
    \vspace{-1em}
 \label{fig:ex_observations}
\end{figure}

\vspace{\vsp}
\paragraph{Fluorescent+LED case}
When light with spiky spectral peaks, such as fluorescent light, passes through the diffraction grating, the resulting diffracted observation exhibits a similar spiky pattern, as shown in \fref{fig:ex_observations}. Therefore, the fluorescent light provides a strong cue for determining the pixel-to-wavelength mapping.
In this case, we use a similar approach to~\cite{karge2018using} that associates corresponding peak positions in the pixel and wavelength domains.
Since the spectral sensitivity $\V{s}$ is unknown, we weigh the illumination spectra $\V{e}$ by the mean spectral sensitivity $\bar{\V{s}}$ as $\bar{\V{s}}\odot\V{e}$, where $\bar{\V{s}}$ is obtained from the same dataset used for the camera spectral sensitivity basis \cite{solomatov2023spectral}, and $\odot$ represents an element-wise multiplication operator. We have observed that the unknown grating efficiency $\boldsymbol{\eta}$ is negligible in determining the peak positions due to its low-frequency distribution.

We select the top two peaks from each of the pixel and wavelength domains for each channel to determine the polynomial parameters $a$, $b$, and $c$ of the pixel-to-wavelength mapping $g$ via linear least squares. Once the pixel-to-wavelength mapping $g$ and weight matrix $\V{W}$ are obtained, we take another capture using an LED light for obtaining the direct and diffracted measurements used in \sref{sec:camspecs-est}.




\vspace{\vsp}
\paragraph{LED only case}
To simplify the recording process, it is preferable if we can work only with an LED.
In this case, due to the low-frequency incoming spectra without obvious peaks, we cannot take the strategy of relating peak positions used in the Fluorescent+LED case.

Therefore, instead of finding peak positions, we frame the problem as finding the optimal horizontal mapping (pixel-to-wavelength) between two plots---the normalized diffraction observations $\mathbf{u}$ and the normalized mean camera spectral sensitivity $\mathbf{v}$ that is scaled by illumination spectrum, expressed as:
 \begin{equation*}
 \begin{aligned}
        \V{v} &\triangleq \nu\left(\V{e}\odot\bar{\V{s}}\right) &&=
 \begin{bmatrix}
     v(\lambda_1), & \ldots, & v(\lambda_f)
 \end{bmatrix}\in\doubleRp^f \\
        \V{u} & \triangleq\nu\left(\V{m}_\mathrm{dif}\right)&& = 
  \begin{bmatrix}
     u(p_1), & \ldots, & u(p_n)
 \end{bmatrix}
 \in\doubleRp^n,
 \end{aligned}
 \end{equation*}
 where $\nu$ is a vector normalization as $\nu\left(\V{x}\right) = \V{x} / \norm{\V{x}}_2$.

We solve the point-to-plane iterative closest point (ICP) problem~\cite{chen1992object} to determine the optimal polynomial parameters $a$, $b$, and $c$ along the horizontal axis by minimizing the following objective function:
\begin{equation*}
\small
    \begin{aligned}
        \label{eq:icp-objective}
        \mathcal{L}_\mathrm{ICP} &= \sum_{i=1}^n \sum_{j\in\mathcal{N}(i)}\left( 
            \left(
                \begin{bmatrix}
                    a p_i^2 + bp_i + c\\
                    u(p_i) 
                \end{bmatrix}
                - \begin{bmatrix}
                    \lambda_j \\
                     v(\lambda_j)
                \end{bmatrix}
            \right)\transp
            \V{n}_\mathrm{v}(j)
             \right)^2, \\
    \end{aligned}
\end{equation*} 
where $\mathcal{N}(i)$ is the set of indices for the current neighboring points of $\begin{bmatrix} a p_i^2 + bp_i + c, &  u(p_i)  \end{bmatrix}\transp$, and 
$\V{n}_\mathrm{v}(j)\in\mathbb{S} \subset \doubleR^2$
is a unit normal vector to $\left[\lambda_j, {v}(\lambda_j)\right]\transp$ which satisfies 
\begin{equation}
    {\begin{bmatrix}
                    \lambda_j - \lambda_{j+1} \\
                    {v}(\lambda_j) -  {v}(\lambda_{j+1}) 
                \end{bmatrix}
                }\transp
                \V{n}_\mathrm{v}(j) = 0.
\end{equation}
We iteratively minimize the objective function for $500$ iterations, 
with the initial guesses set to $a=0$, $b=f/n$, $c=0$, where $b$ is given as the ratio of the numbers of elements of $\V{u}$ and $\V{v}$, to determine the pixel-to-wavelength mapping $g$. This setting allows for a convenient one-time capture for both determining the pixel-to-wavelength mapping and obtaining the direct and diffracted measurements used in \sref{sec:camspecs-est}.

\newcommand{\paragra}[1]{\noindent \textbf{#1: }}
\section{Experiments}

\label{sec:results}
In this section, we evaluate our method on both synthetic and real-world scenes. We first describe our experimental setup and then present the evaluation results.


\vspace{\vsp}
\paragraph{Baselines} 
We compare the proposed approach with the Exif-based method~(\exif)~\cite{solomatov2023spectral} and the color chart-based method~(\cc)~\cite{jiang2013space}.
We present two variants of the proposed method as detailed in \sref{sec:pix2wave}: Ours~(LED+Flu), which employs a fluorescent lamp observation for pixel-to-wavelength mapping estimation, and Ours~(LED), which uses only an LED light observation.
For the color chart-based method, we use $2$ basis functions, as mentioned in the original paper, which yielded the best results in our preliminary experiments. Since the Exif-based method cannot estimate white balance and introduces a 3$\times$3 ambiguity, we use the color checker observation to resolve this ambiguity. 

We also compared ours with the existing diffraction grating-based approach by Toivonen~\etal~\cite{toivonen2020practical}, adjusted to our experimental setup. However, their method produced unstable estimations due to the limited observations available in our setup compared to their original conditions. Further details are provided in the supplemental material.



\vspace{\vsp}
\paragraph{Basis representations}
To compute the basis for camera spectral sensitivity $s(\lambda)$, we use the dataset from the Exif-based method~\cite{solomatov2023spectral}, which includes $44$ cameras for training and $5$ for testing. 
Following the detailed analysis on basis selection for a camera spectral sensitivity in~\cite{jiang2013space}, we apply SVD to the sensitivities in the training dataset in a per-channel manner. For the grating efficiency, we use a Fourier basis. Further discussion on the grating efficiency basis can be found in the supplementary material.
The spectral resolution~$f$ in the dataset of camera spectral sensitivities is~$f=31$ from \SI{400}{\nano\meter} to \SI{700}{\nano\meter}. Assuming the number of basis for the camera spectral sensitivity and the grating efficiency to be the same $b_\eta = b_s,$ the required condition for the number of bases for our method is $b_\eta=b_s\le 23$ based on the discussion in \sref{sec:solution}.
For robust estimation, the number of basis functions is set to $7$ for each channel of the camera spectral sensitivity and the grating efficiency across all experiments.
%


\vspace{\vsp}
\paragraph{Evaluation metrics}
Following~\cite{solomatov2023spectral}, we use the average relative full-scale error across the three color channels, $\mathrm{RE}$, which is defined as:
\begin{equation}
\label{eq:metric}
{\small
\begin{aligned}
    \mathrm{RE} &= \frac{1}{3}\sum_{c}
    \frac{\mathrm{RMSE}(c)}{\max({\V{s}_c})}, ~    
    \mathrm{RMSE}(c) &= \sqrt{\frac{\norm{\V{s}^*_c - \V{s}_c}_2^2}{f}},
\end{aligned}
}
\end{equation}
where $\V{s}^*_c$ and $\V{s}_c$ denote the estimated and ground-truth camera spectral sensitivity for each channel $c\in\{R,G,B\}$.

\subsection{Evaluation with synthetic scenes}
\begin{figure*}[t]
\begin{minipage}[t]{\linewidth}
\centering
\small
\begin{tabular}{@{}c@{\hspace{0.3em}}c@{}c@{}c@{}c@{}c|c@{}c@{}c@{}c@{}}
\makebox[0.005000\linewidth]{} 
& 
\makebox[0.005000\linewidth]{} 
&
\multicolumn{4}{c}{(a) Synthetic experiment}
& 
\multicolumn{4}{c}{(b) Real-world experiment}
\\
\vspace{-0.8em}
 &  &  &  &  &  &  &  &  &    \\
\makebox[0.005000\linewidth]{} & 
\makebox[0.005000\linewidth]{} &
\makebox[0.12000\linewidth]{EOS 650D} &
\makebox[0.12000\linewidth]{Olympus EPL2} &
\makebox[0.12000\linewidth]{Pentax K5} &
\makebox[0.12000\linewidth]{Galaxy S20} & 
\makebox[0.12000\linewidth]{EOS RP} &
\makebox[0.12000\linewidth]{iPhone 15ProMax} &
\makebox[0.12000\linewidth]{\sony} &
\makebox[0.12000\linewidth]{\djipocket} \\

\raisebox{0.0600\linewidth}{\makebox[0.005000\linewidth]{\rotatebox[origin=c]{90}{Ours}}} & 
\raisebox{0.0600\linewidth}{\makebox[0.005000\linewidth]{\rotatebox[origin=c]{90}{(LED+Flu)}}} & 
\includegraphics[width=0.12000\linewidth]{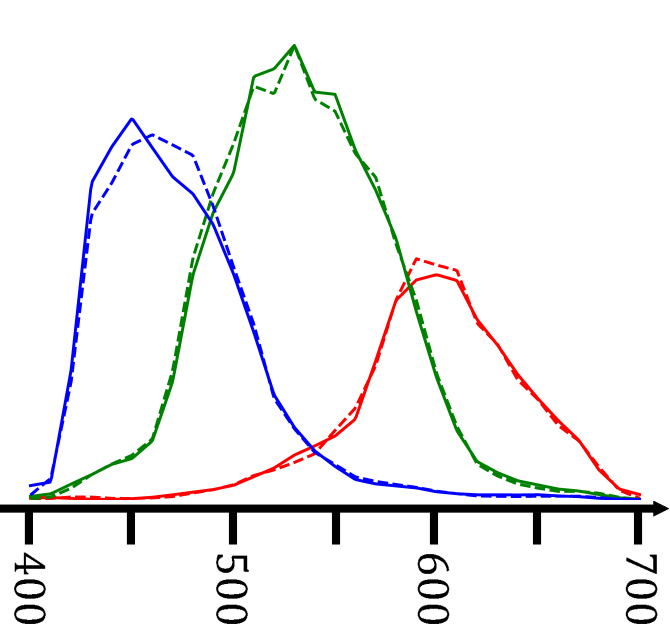} &
\includegraphics[width=0.12000\linewidth]{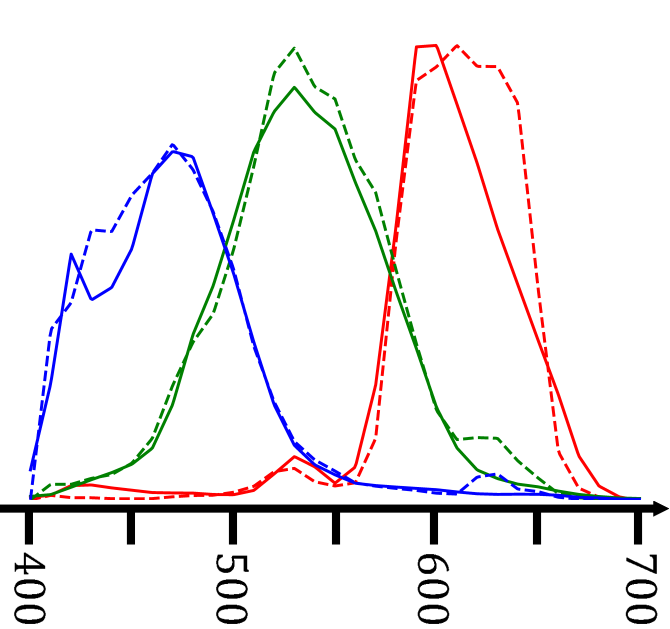} &
\includegraphics[width=0.12000\linewidth]{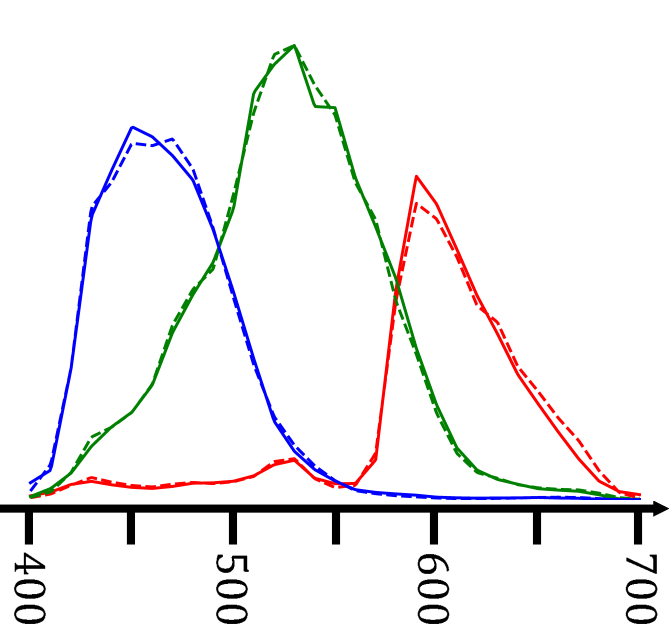} &
\includegraphics[width=0.12000\linewidth]{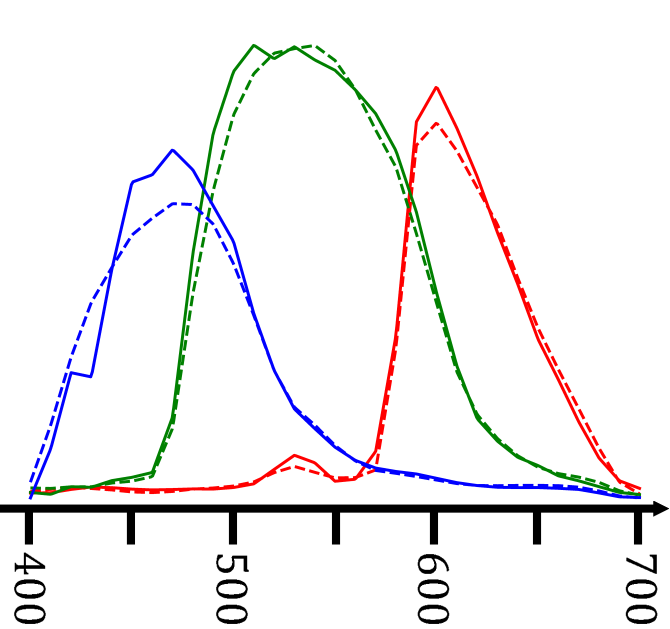} & 
\includegraphics[width=0.12000\linewidth]{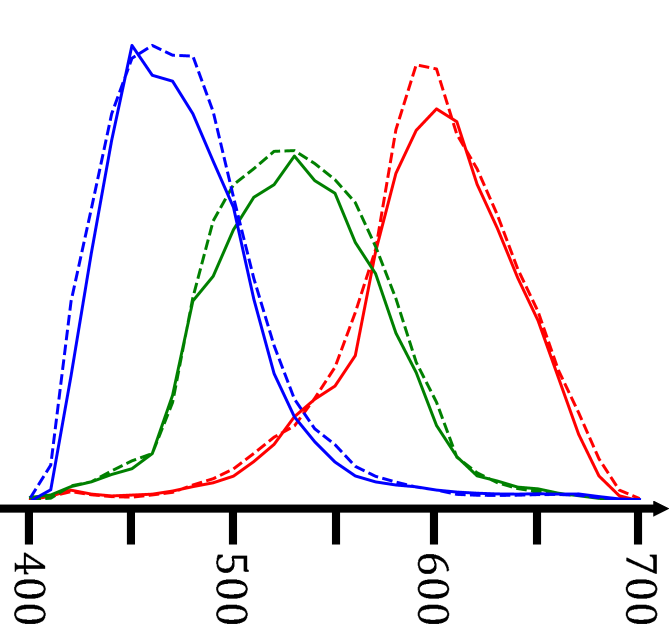} &
\includegraphics[width=0.12000\linewidth]{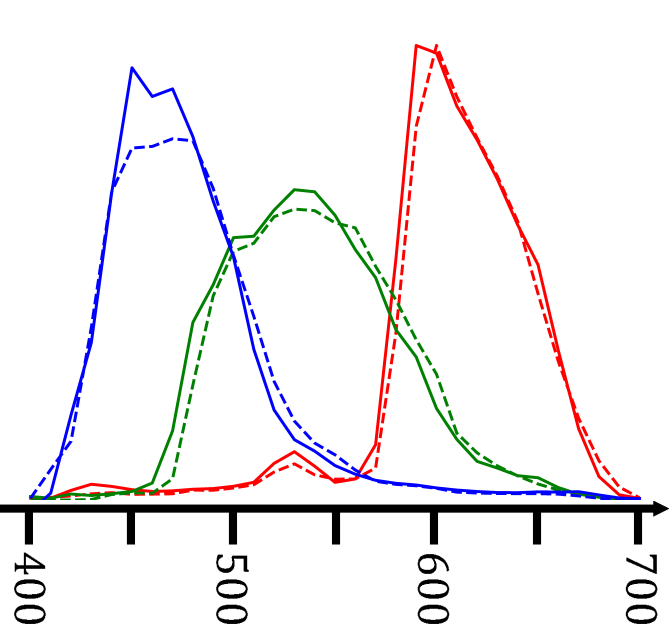} &
\includegraphics[width=0.12000\linewidth]{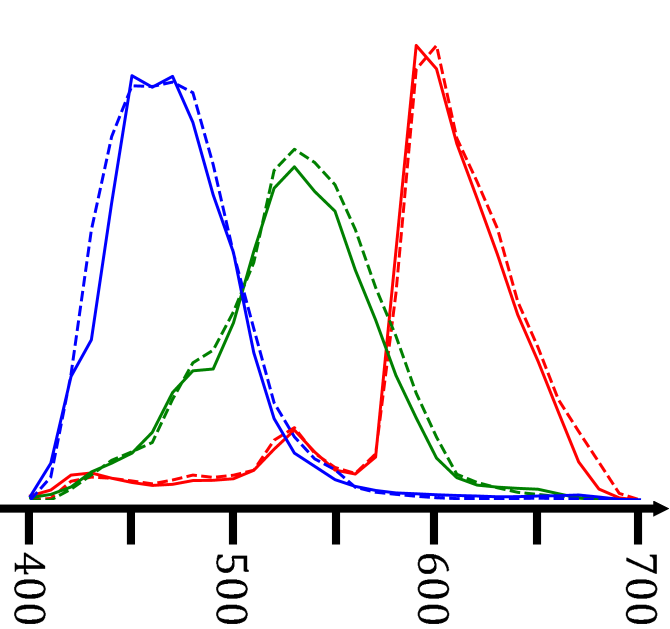} &
\includegraphics[width=0.12000\linewidth]{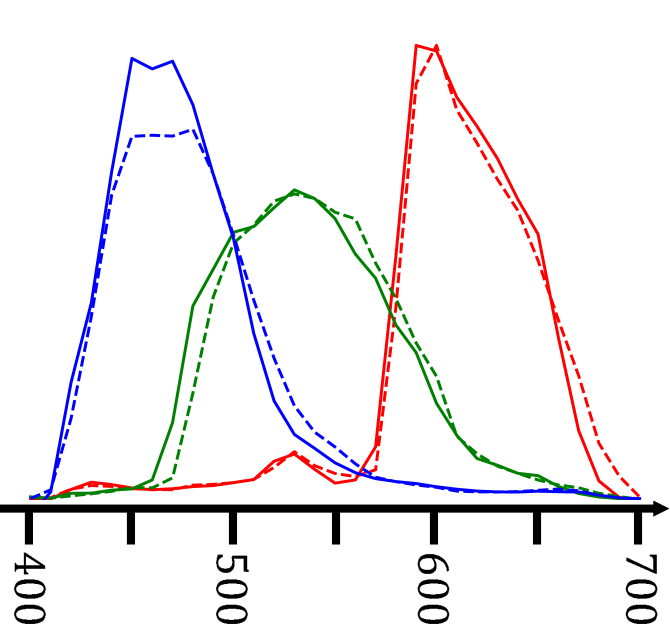} 
\\
&
& \bestscore{0.12000\linewidth}{$\mathbf{2.84}$} & \bestscore{0.12000\linewidth}{$\mathbf{7.25}$} & \bestscore{0.12000\linewidth}{$\mathbf{2.17}$} & \bestscore{0.12000\linewidth}{$\mathbf{4.16}$} 
& \bestscore{0.12000\linewidth}{$\mathbf{3.53}$} & \secondbest{0.12000\linewidth}{$\num{5.36}$} & \bestscore{0.12000\linewidth}{$\mathbf{4.17}$} & \secondbest{0.12000\linewidth}{$\num{5.77}$} \\


\multicolumn{2}{c}{\raisebox{0.0600\linewidth}{\makebox[0.005000\linewidth]{\rotatebox[origin=c]{90}{Ours (LED)}}}} & 
\includegraphics[width=0.12000\linewidth]{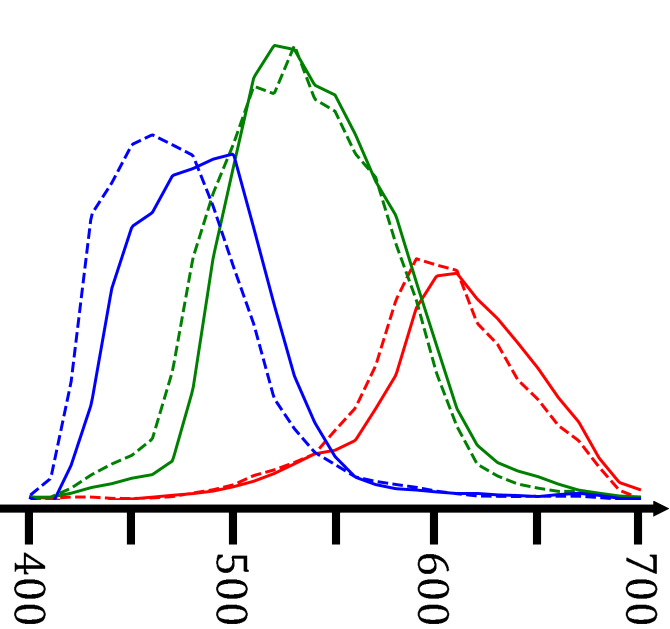} &
\includegraphics[width=0.12000\linewidth]{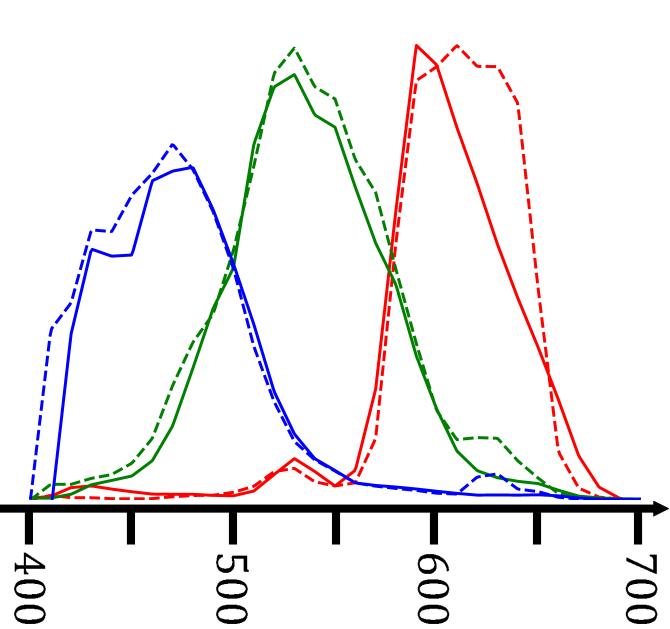} &
\includegraphics[width=0.12000\linewidth]{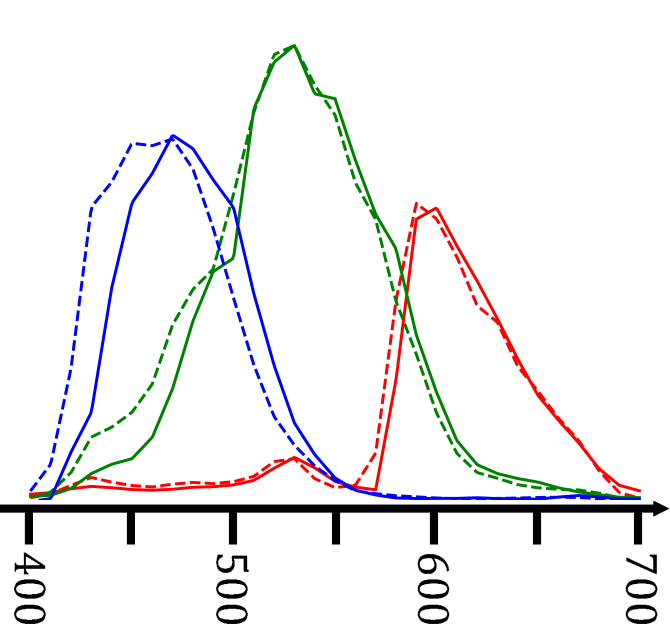} &
\includegraphics[width=0.12000\linewidth]{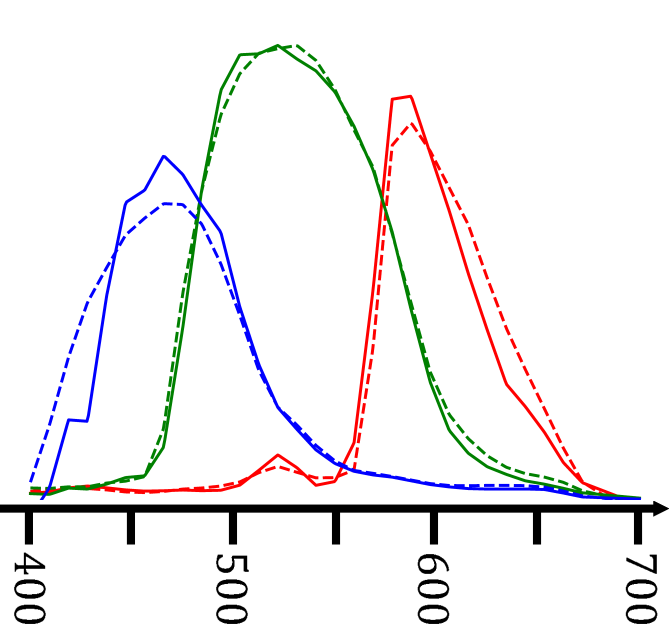} & 

\includegraphics[width=0.12000\linewidth]{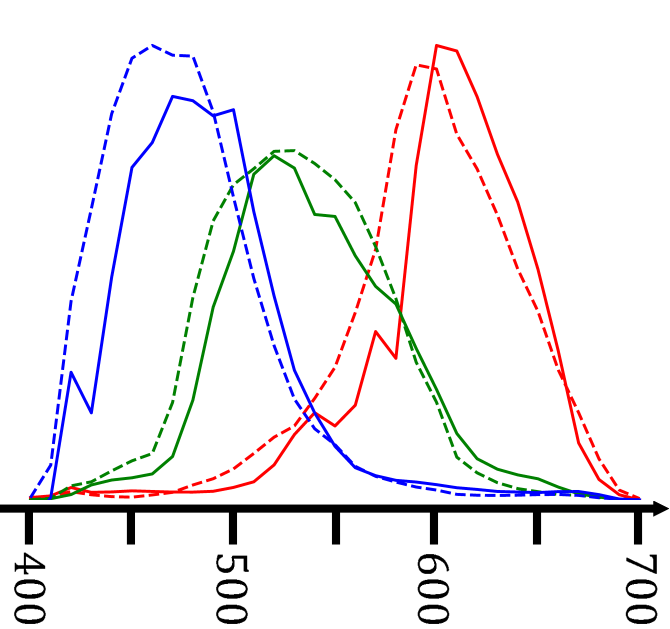} &
\includegraphics[width=0.12000\linewidth]{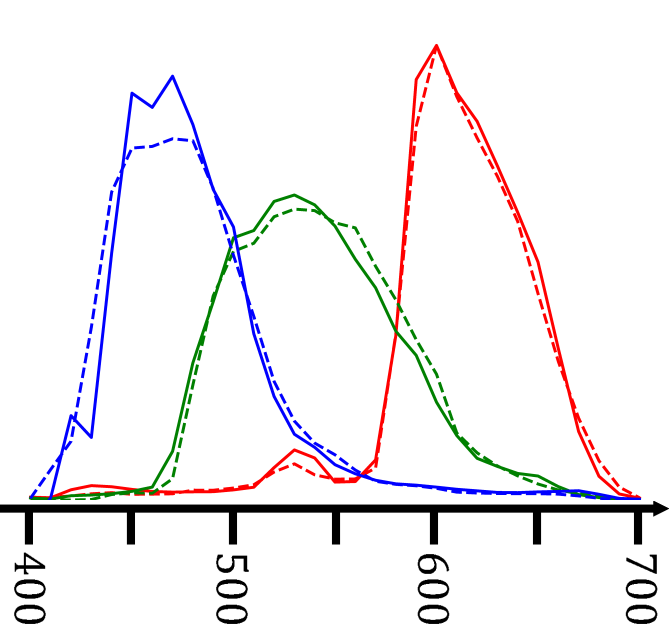} &
\includegraphics[width=0.12000\linewidth]{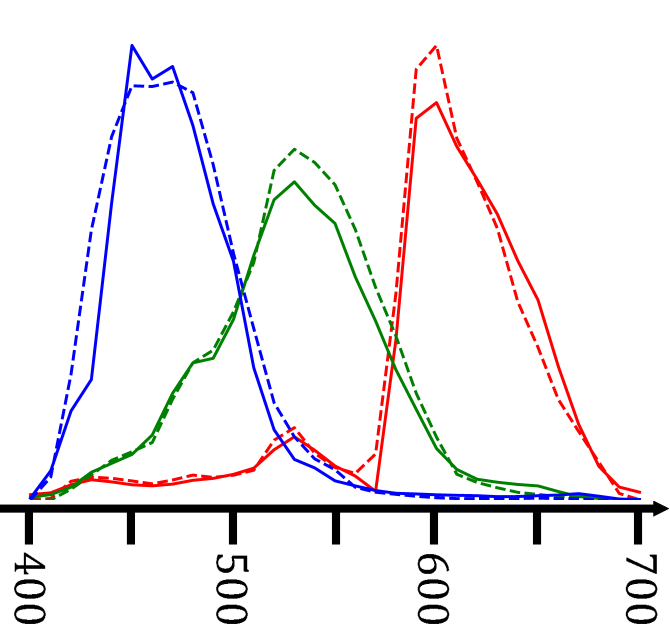} &
\includegraphics[width=0.12000\linewidth]{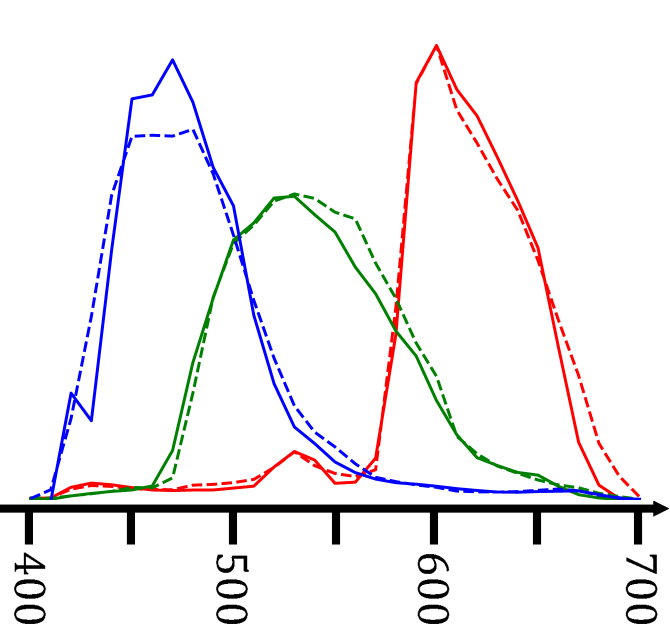} 
\\
& & $\num{11.2}$ & {$\num{8.81}$} & $\num{8.81}$ & {$\num{6.47}$} &  {$\num{11.9}$} & \bestscore{0.12000\linewidth}{$\mathbf{5.12}$} & \secondbest{0.12000\linewidth}{$\num{5.45}$} & \bestscore{0.12000\linewidth}{$\mathbf{5.76}$} \\
\multicolumn{2}{c}{\raisebox{0.0600\linewidth}{\makebox[0.005000\linewidth]{\rotatebox[origin=c]{90}{Exif+CC~\cite{solomatov2023spectral}}}}} & 
\includegraphics[width=0.12000\linewidth]{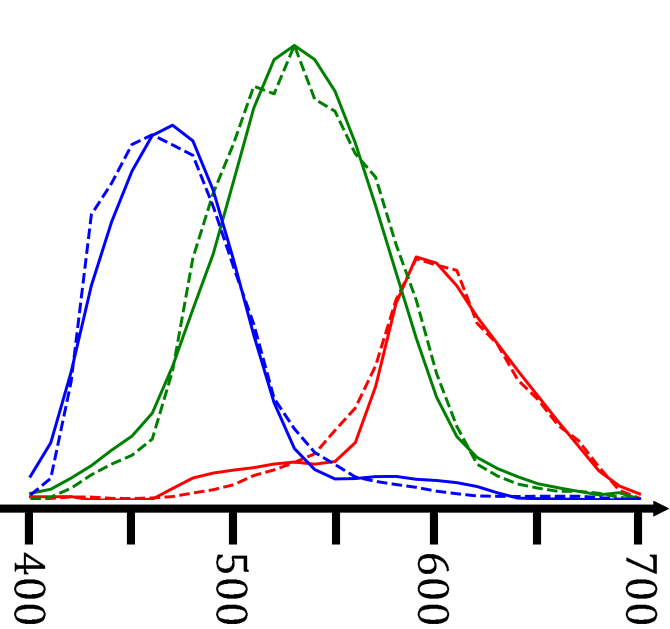} &
\includegraphics[width=0.12000\linewidth]{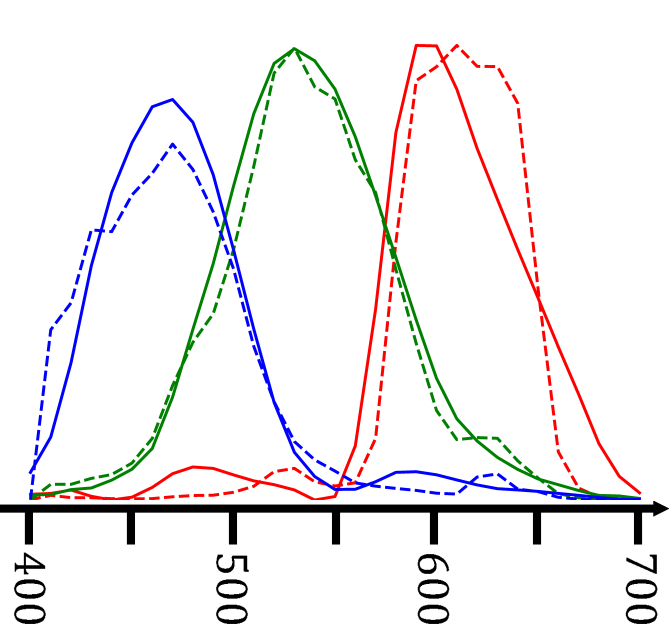} &
\includegraphics[width=0.12000\linewidth]{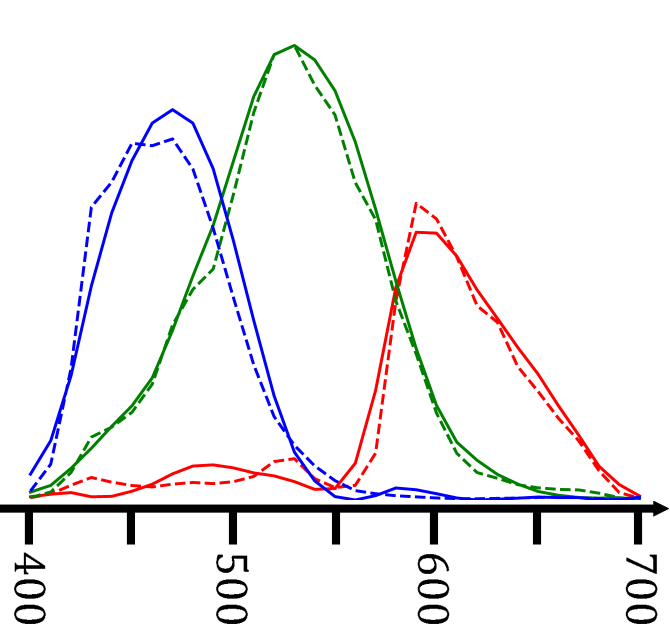} &
\includegraphics[width=0.12000\linewidth]{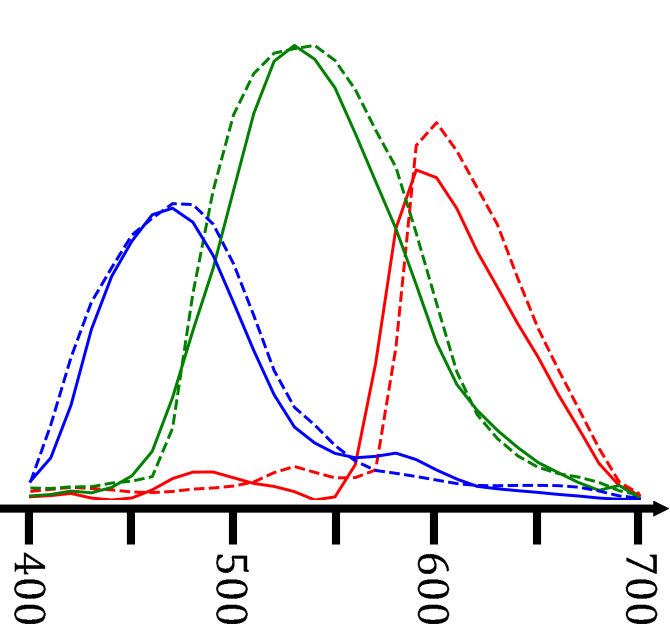} &
\includegraphics[width=0.12000\linewidth]{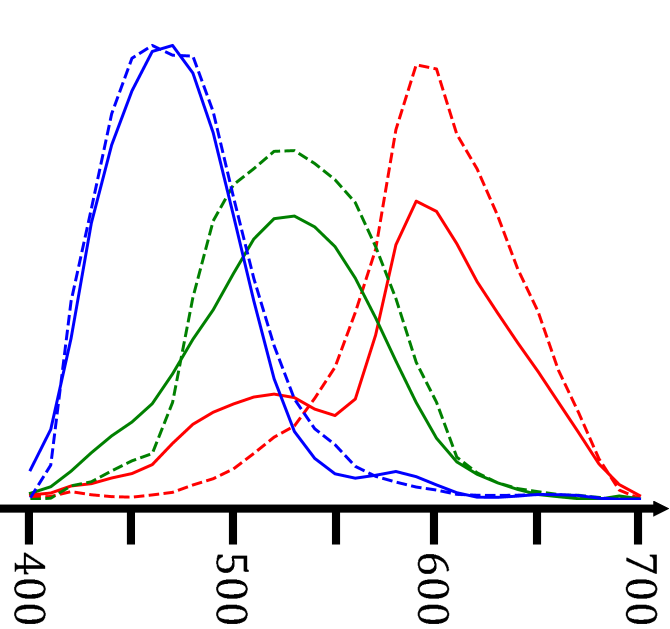} &
\includegraphics[width=0.12000\linewidth]{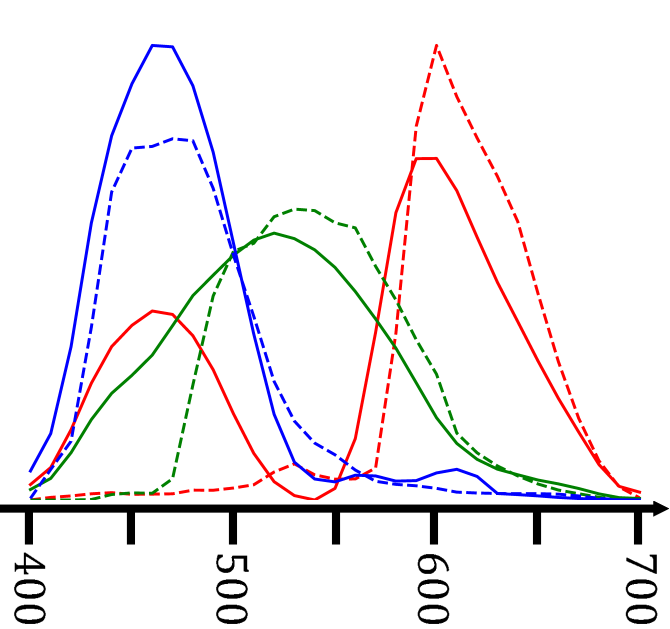} &
\includegraphics[width=0.12000\linewidth]{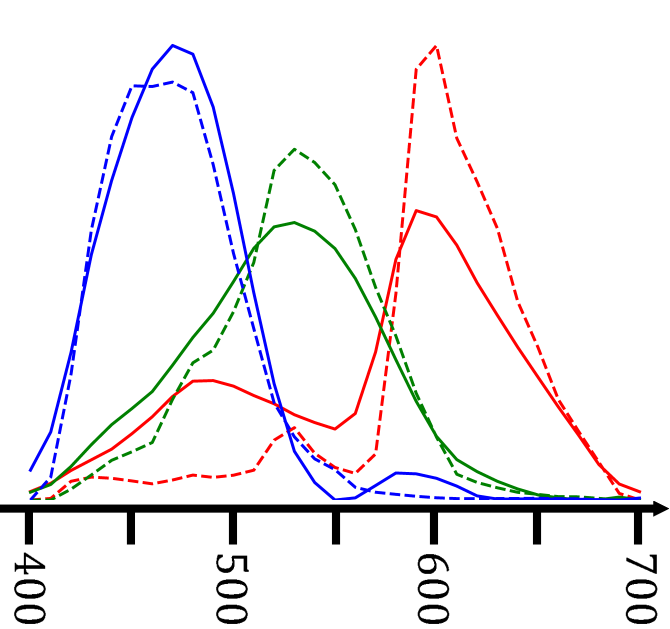} &
\includegraphics[width=0.12000\linewidth]{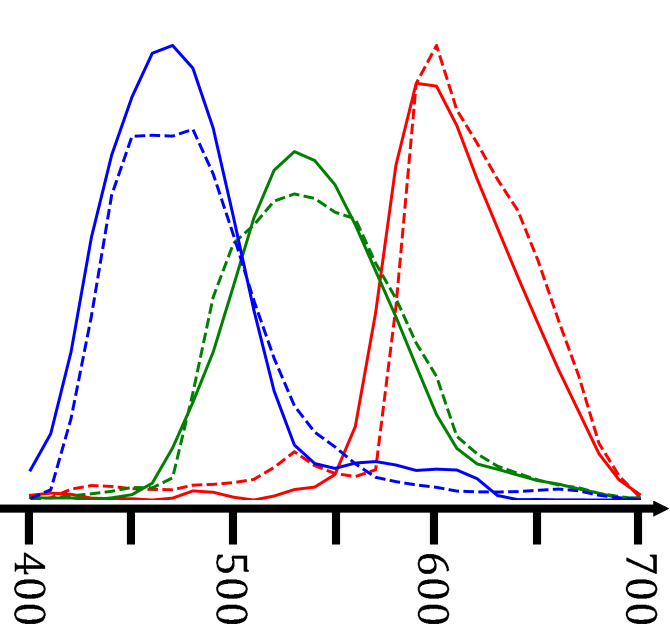} 
\\
& & {$\num{5.02}$} & $\num{8.56}$ & {$\num{5.02}$} & $\num{6.89}$ & {$\num{8.18}$} & $\num{15.0}$ & $\num{9.45}$ & $\num{7.68}$ \\

\multicolumn{2}{c}{\raisebox{0.0600\linewidth}{\makebox[0.005000\linewidth]{\rotatebox[origin=c]{90}{CC~\cite{jiang2013space}}}}} & 
\includegraphics[width=0.12000\linewidth]{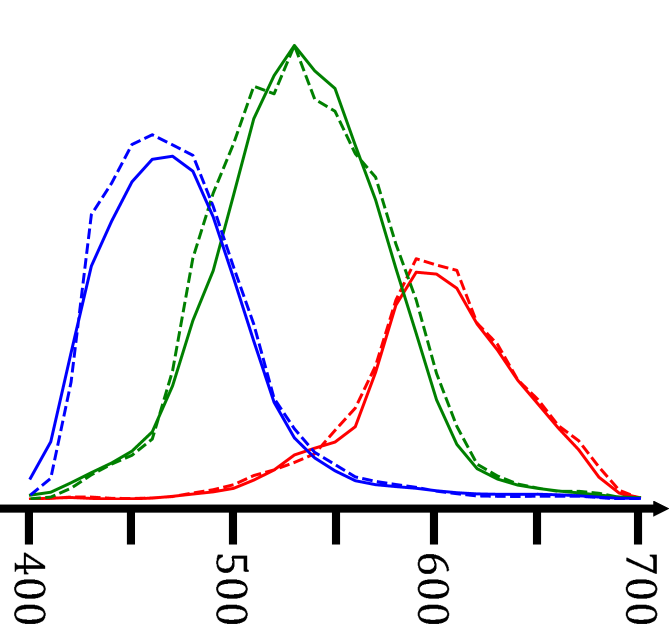} &
\includegraphics[width=0.12000\linewidth]{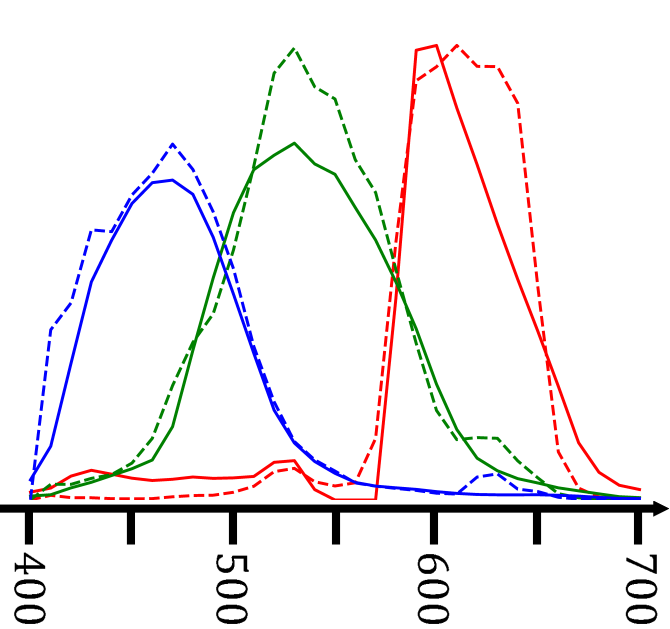} &
\includegraphics[width=0.12000\linewidth]{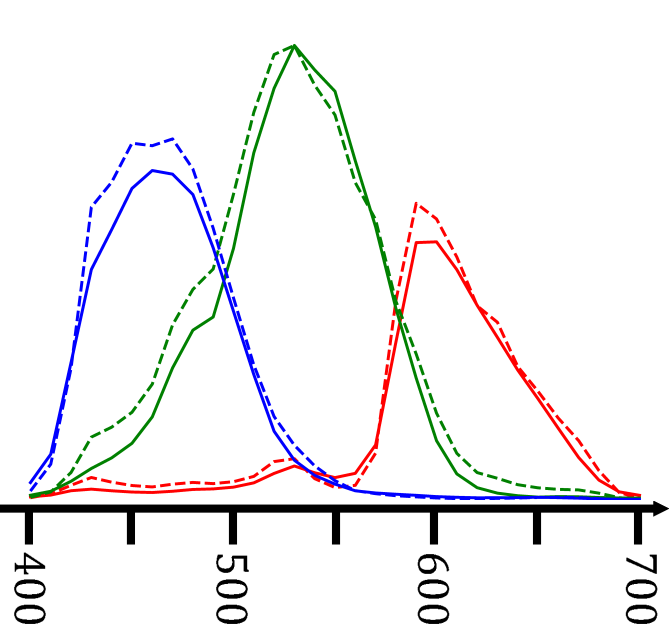} &
\includegraphics[width=0.12000\linewidth]{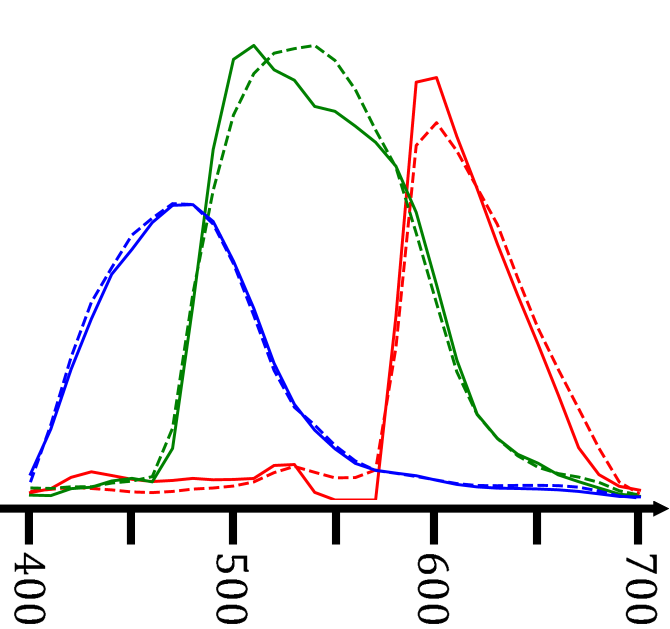} & 

\includegraphics[width=0.12000\linewidth]{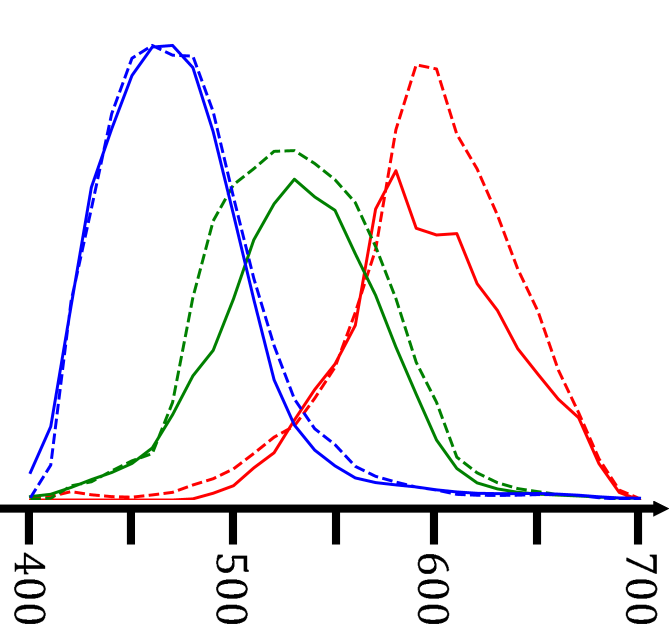} &
\includegraphics[width=0.12000\linewidth]{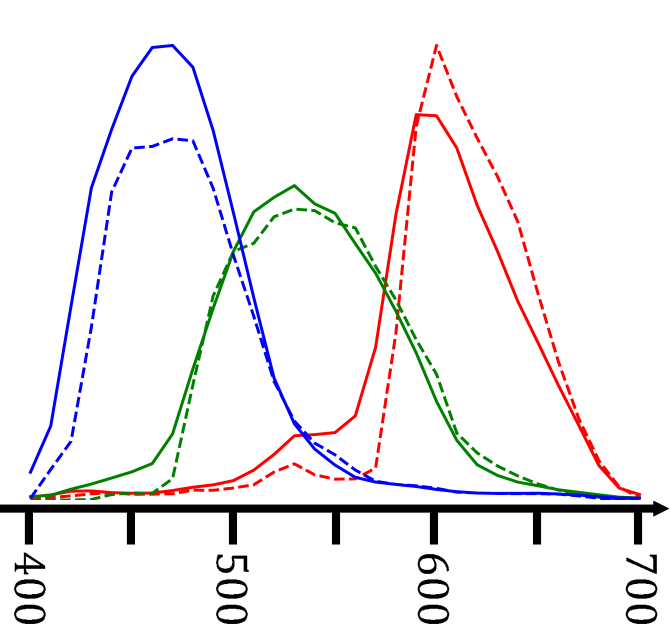} &
\includegraphics[width=0.12000\linewidth]{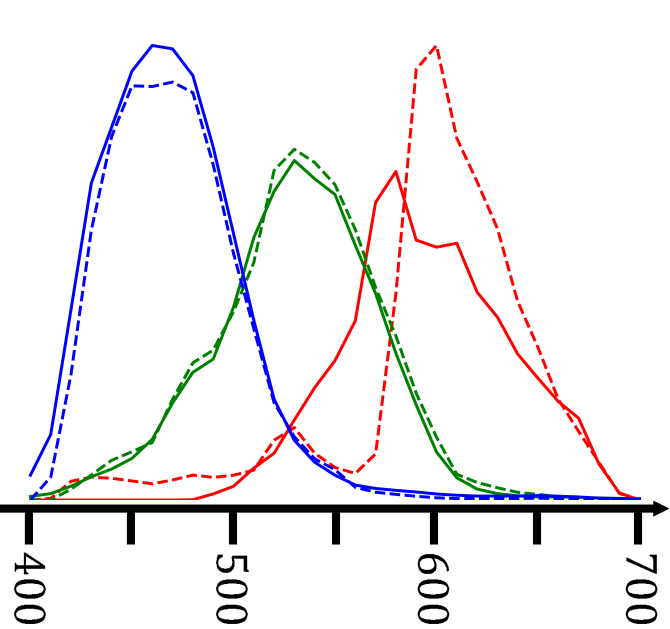} &
\includegraphics[width=0.12000\linewidth]{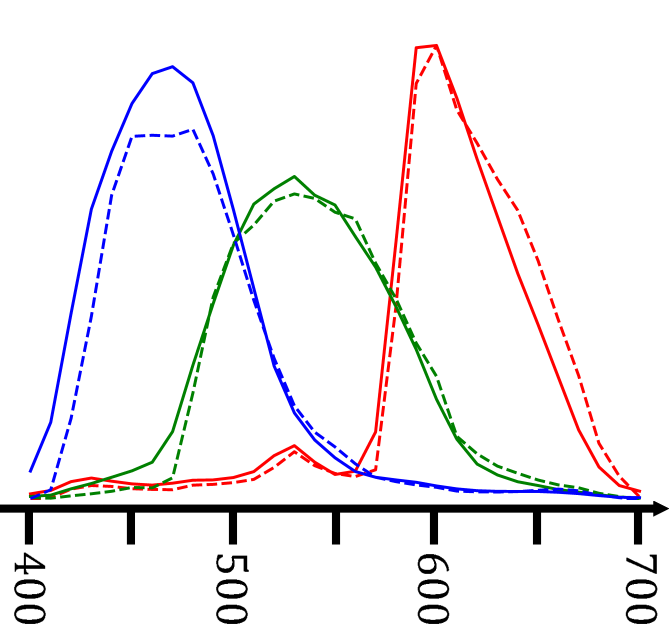} \\
& & \secondbest{0.12000\linewidth}{$\num{3.75}$} & \secondbest{0.12000\linewidth}{$\num{8.04}$} & \secondbest{0.12000\linewidth}{$\num{3.94}$} & \secondbest{0.12000\linewidth}{$\num{4.25}$} 
& \secondbest{0.12000\linewidth}{$\num{8.45}$} 
& $\num{9.13}$ & $\num{8.99}$ & $\num{6.59}$ \\

\end{tabular}
\end{minipage}\\
\makebox[0.7\linewidth]{} 
\begin{minipage}[r]{0.300\linewidth}
\centering
\includegraphics[width=\linewidth]{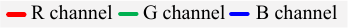} 
\end{minipage}
    \caption{
    Results of the proposed and comparison methods for our (a) synthetic and (b) real-world datasets. For each setting and method, we show plots of the estimated (solid line) and ground-truth (dotted line) spectral sensitivity for each color channel. 
    All plots are normalized by the maximum response across all channels.
    The values below the plots indicate the $\mathrm{RE}$ defined in~\eref{eq:metric}. 
    For simplicity, the factor of~$10^{-2}$ is omitted. Underscored bold and underscored score are the best and second-best result for each setting, respectively. 
    }
    \label{fig:realxsynth-exp}
\end{figure*}

In this section, we quantitatively evaluate the comparison methods and ours on a synthetic dataset. 

\vspace{\vsp}
\paragraph{Dataset}
We render the direct and diffracted images using \eref{eq:1st-image-formation}, with parameters as follows. We use five camera spectral sensitivities: \syneos, \syniphone, \synolympus, \synpentax, and \synsumsung, all from the same test dataset as the Exif-based method.
The D65 illuminant, as defined by the CIE Standard Illuminants, is used as the incoming light.
To synthesize the pixel-to-wavelength mapping, we randomly define the mapping while ensuring that the visible wavelength range is fully covered in the diffraction image. 
The grating efficiency function is set to a low-frequency response as illustrated in \fref{fig:method}, parameterized by $7$ Fourier bases.
Since the Exif-based method relies on metadata rather than images, we evaluate their method using real-world data.

\vspace{\vsp}
\paragraph{Results}
\Fref{fig:realxsynth-exp}(a) shows the estimated results of the proposed and comparison methods for our synthetic scenes. 
For ideal observations, both Ours (LED+Flu) and the color chart-based method, \cc, yield low estimation errors, as both methods are guaranteed to produce optimal results under the basis representation assumption. The remaining errors stem from the gap between the assumption and the synthetic data.
The Exif-based method, \exif, produces a generally reasonable overall shape, but it fails to accurately capture finer details.
Ours~(LED) yields a larger error than Ours~(LED+Flu) due to estimation errors in the pixel-to-wavelength mapping when using only LED observations; however, overall, we can still obtain a reasonable estimation even by capturing just a single calibration scene.


\subsection{Evaluation with real-world scenes}
Here, we present evaluations using our real-world dataset.

\vspace{\vsp}
\paragraph{Dataset}
\Fref{fig:setup-and-observation} illustrates our experimental setup, which consists of a light source, a diffraction grating sheet, and a camera. We capture the images in a dark room for both our method and the comparison methods, ensuring that the only light in the scene comes from illumination with known spectra (LED or fluorescent).
We use a 3D-printed slit for the incoming light, positioned in front of the light source, and place a diffraction grating in front of the camera, manually aligned roughly parallel to the image plane. 
For the spiky light source, we use a Pa-Look Ball~(EFD15EL)\footnote{Panasonic Corporation} fluorescent lamp.
We use the following four cameras: \eos, \iphonexv, \sony, and \djipocket. 
For the non-spiky light sources and diffraction sheets, we use three different products for each.
In this section, we present the results for one setup. Additional experiments with other setups are provided in the supplementary material.

Due to the high dynamic range of the direct and diffracted light, we capture HDR images by taking two images with different exposure times for each. 
The extra diffracted image under fluorescent light is captured for LED+Flu, solely for pixel-to-wavelength mapping. 
After capturing the direct and diffracted light observations, we roughly crop them with a sufficient margin at both ends of the diffraction pattern along the diffraction direction. 
The diffracted light observation is cropped to approximately $200\times1000~\mathrm{pix}$ and aggregated to $1\times1000~\mathrm{pix}$.

For the comparison methods, we use a 24-patch color checker chart\footnote{ColorChecker Classic, Calibrite} with spectral reflectance data from Chromaxion\footnote{Chromaxion, \url{https://www.chromaxion.com}
, last accessed on March 7, 2025.
}.
The color checker chart is captured under the same non-spiky light source used in our experiments. We carefully position the color checker chart so that it is parallel to the camera and uniformly illuminated by the light source, ensuring that no shadows are cast.
The metadata for the Exif-based method is extracted from our captured data following the procedure described in their paper~\cite{solomatov2023spectral}. 

The ground-truth spectral sensitivities are obtained using a calibration device\footnote{camSPECS XL, Image Engineering GmbH \& Co. KG} with a lens attached. The illuminant spectra of the light sources, shown in \fref{fig:ex_observations}, are measured through the diffraction grating using a spectrometer\footnote{CS-2000S, KONICA MINOLTA, INC.}.

\vspace{\vsp}
\paragraph{Results}
\Fref{fig:realxsynth-exp}(b) presents the estimated results of the proposed and comparison methods. 
Overall, the proposed methods achieve the lowest estimation errors for most scenes, and Ours (LED) produces results nearly comparable to Ours (LED+Flu). 
For the \eos scene, Ours~(LED) yields a larger error due to the failure of pixel-to-wavelength mapping estimation using a non-spiky light source.
The color checker-based method, \cc, is relatively sensitive to real-world noise due to the low dimensionality of the spectral reflectances in the color patches, as discussed in previous works~\cite{toivonen2020practical, dicarlo2004emissive}.
While the Exif-based method, \exif, has an advantage of being easy to use, as it only relies on metadata for recovering sensitivity, it yields larger estimation errors across all scenes.





\section{Conclusions}
We present a method for estimating camera spectral sensitivity using an uncalibrated diffraction grating sheet. Compared to previous methods that use reference targets, such as a color checker chart, our approach achieves highly accurate estimation by leveraging the diffraction grating.
Unlike previous diffraction grating-based methods, the proposed method jointly estimates both camera spectral sensitivity and grating efficiency without needing additional recordings, making it practical, cost-efficient, and easy-to-use.

One of the limitations of our method is that pixel-to-wavelength mapping estimation breaks down if the illumination has uniform spectra. In this case, the diffraction image simply shows the camera spectral sensitivity scaled by unknown grating efficiency, providing no clues for the mapping estimation. However, in practice, readily-available light sources do not have such uniform spectra. LEDs, for example, tend to have a non-uniform spectral distribution, enabling us to estimate the mapping and camera spectral sensitivities, as demonstrated in our real-world experiments.
\section*{Acknowledgments}
This work was supported by JSPS KAKENHI Grant Numbers JP23H05491 and JP25K00142, JST ASPIRE Grant Number JPMJAP2304, the Canada First Research Excellence Fund program (VISTA), an NSERC Discovery Grant, and the Canada Research Chair program.

{
    \small
    \bibliographystyle{ieeenat_fullname}
    \bibliography{ref}
}

\end{document}


\maketitle

\noindent In this supplementary material, 
\begin{itemize}
\item[A.] we provide a detailed description of the experimental setup.
\item[B.] we present additional experimental results using different light sources and diffraction grating sheets.
\item[C.] we present additional experimental results of pixel-to-wavelength mapping of our method.
\item[D.] we provide a further derivation of the pixel-to-wavelength mapping representation.
\item[E.] we provide a detailed description of the basis representations for grating efficiency. 
\item[F.] we provide a detailed description of the linear system in the proposed solution method.
\item[G.] we provide an additional comparison with the existing diffraction grating-based method.
\item[H.] we present a detailed conditions for one-to-one pixel-to-wavelength mapping to be hold.
\item[I.] we evaluate the consistency of grating efficiency estimations across different cameras.
\item[J.] we present the raw measurements captured in the real-world experiments.
\end{itemize}


\begin{figure}
    \centering
    \includegraphics[width=\linewidth]{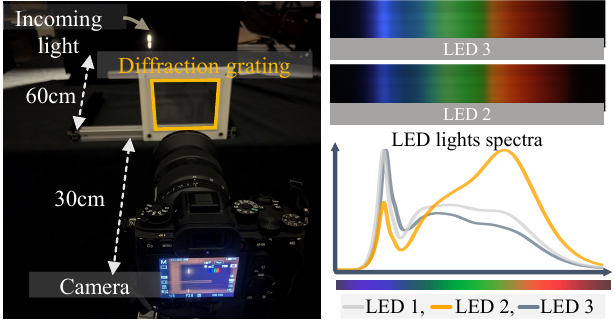}
    \caption{
      The left-hand side image shows our experimental setup with the specific distance between the light source with a slit, the diffraction grating, and the camera. 
      The second row shows the plots of the light spectra of LED~1, LED~2, and LED~3 and examples of diffracted observations. $ x$-axis and $y$-axis correspond to wavelength and the intensity in each wavelength, respectively. The observations are adjusted for visualization. 
    }
    \label{supfig:observation-ex}
\end{figure}
\section{Real-world experimental details}
\Fref{supfig:observation-ex}~(Left) shows the our experimental setup with approximate distances annotated. The incoming light goes through the 3D-printed slit with the \SI{2.5}{\milli\meter} width. 

The images are captured in a completely dark room setting with the RAW format, with all camera parameters manually fixed, including white balance, ISO, and exposure time. 
The white balance is set to daylight and the ISO to $100$ throughout the experiment. 

\section{Experiment results using different light sources and grating sheets}
To verify the effects of different incoming light spectra and differences among diffraction grating sheets, we compare the results using three different light sources and three different diffraction grating sheets.
For the light sources, we use the following three LEDs:
\begin{itemize}[labelwidth=1.5cm, labelsep=0.5em, leftmargin=!]
    \item[LED~1:] ``Godox LITEMONS LED6R (GODOX Photo Equipment Co., Ltd.),'' with a color temperature of $\SI{6500}{\kelvin}$.
    \item[LED~2:] The same product as LED~1, with a color temperature of $\SI{3200}{\kelvin}$. 
    \item[LED~3:] ``Phone Selfie Light~(Mionondi),'' with a color temperature of $\SI{3200}{\kelvin}$.
\end{itemize} 
The light spectra and the examples of the diffracted observations are shown in \fref{supfig:observation-ex}.
For the diffraction grating sheets, we use the following three sheets:
\begin{itemize}[labelwidth=1.5cm, labelsep=0.5em, leftmargin=!]
    \item[Grating~1:] $\SI{500}{slits/\milli\meter}$, ``$500$ lines/mm Linear Diffraction Grating Sheet'' sold by Bartovation.
    \item[Grating~2:] $\SI{1000}{slits/\milli\meter}$, ``$1000$ lines/mm Linear Diffraction Grating Sheet'' sold by Bartovation.
    \item[Grating~3:] $\SI{500}{slits/\milli\meter}$, ``Diffraction Grating Sheet Replica~500'' sold by Kenis.
\end{itemize}
We use LED~1 and Grating~1 in the main paper.

\begin{figure}[t]
        \centering
        \begin{tabular}{@{}c@{}c@{}c@{}}
\makebox[0.333333\linewidth]{LED~1}
&
\makebox[0.333333\linewidth]{LED~2}
&
\makebox[0.333333\linewidth]{LED~3} \\
\includegraphics[width=0.333333\linewidth]{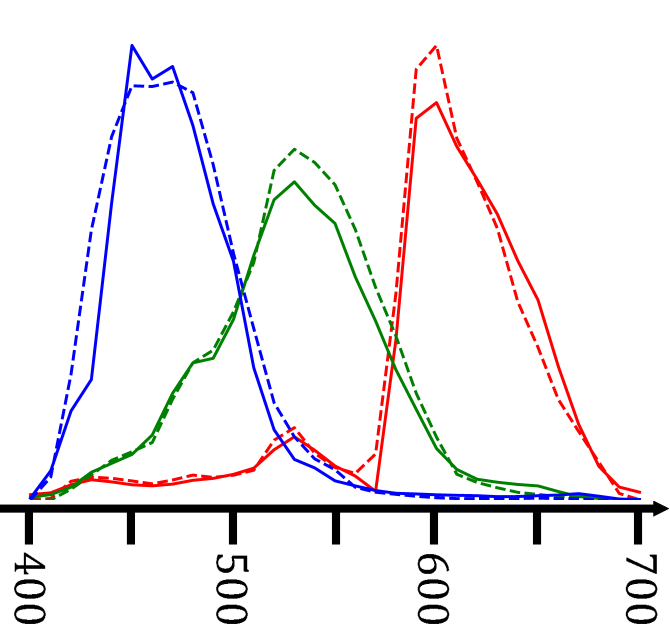}
&
\includegraphics[width=0.333333\linewidth]{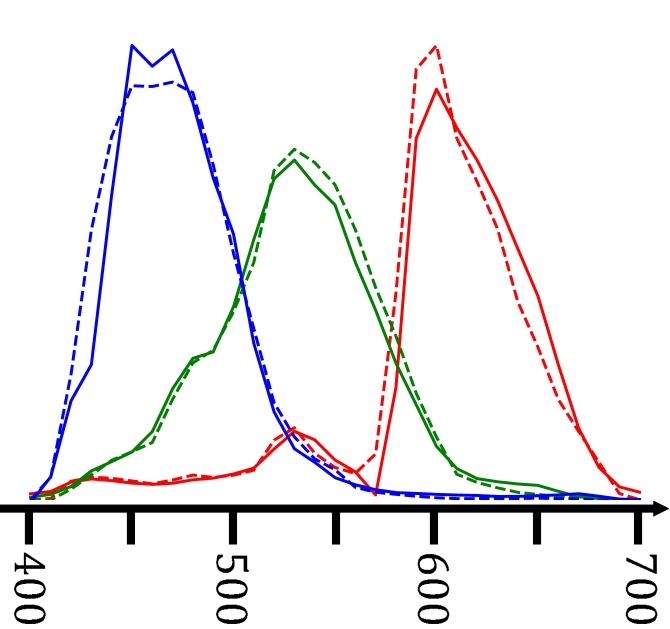}
&
\includegraphics[width=0.333333\linewidth]{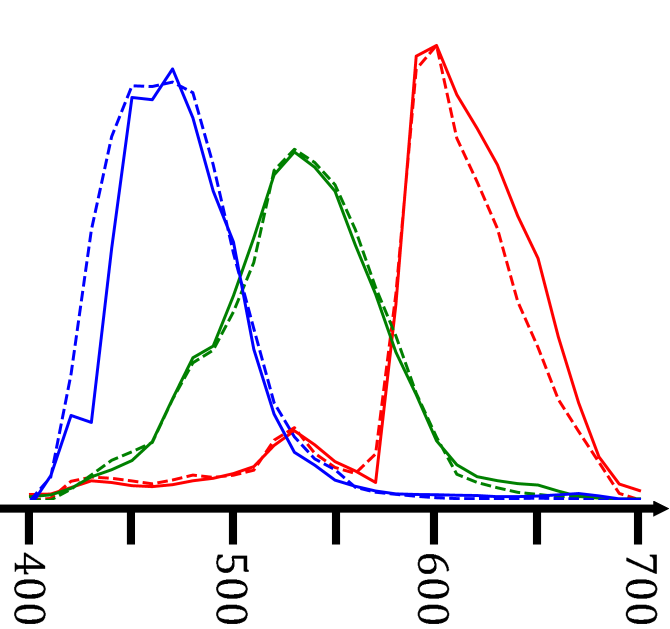}\\
{$\num{5.45E-02}$} & {$\num{4.28E-02}$} & {$\num{6.42E-02}$} \\
        \end{tabular}
\caption{Estimated results of our method using different light sources for \sony with Grating~1}
\label{supfig:real-leds-comparison}
        \end{figure}

\begin{figure}[t]
\begin{tabular}{@{}c@{}c@{}c@{}}
\makebox[0.333333\linewidth]{Grating~1}
&
\makebox[0.333333\linewidth]{Grating~2}
&
\makebox[0.333333\linewidth]{Grating~3} \\
\makebox[0.333333\linewidth]{[\SI{500}{slits\per\milli\meter}]}
&
\makebox[0.333333\linewidth]{[\SI{500}{slits\per\milli\meter}]}
&
\makebox[0.333333\linewidth]{[\SI{1000}{slits\per\milli\meter}]} \\
\includegraphics[width=0.333333\linewidth]{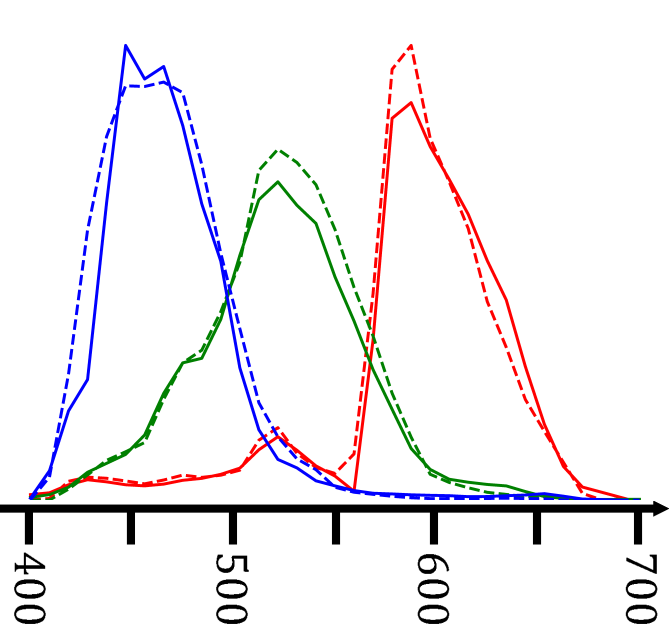}
&
\includegraphics[width=0.333333\linewidth]{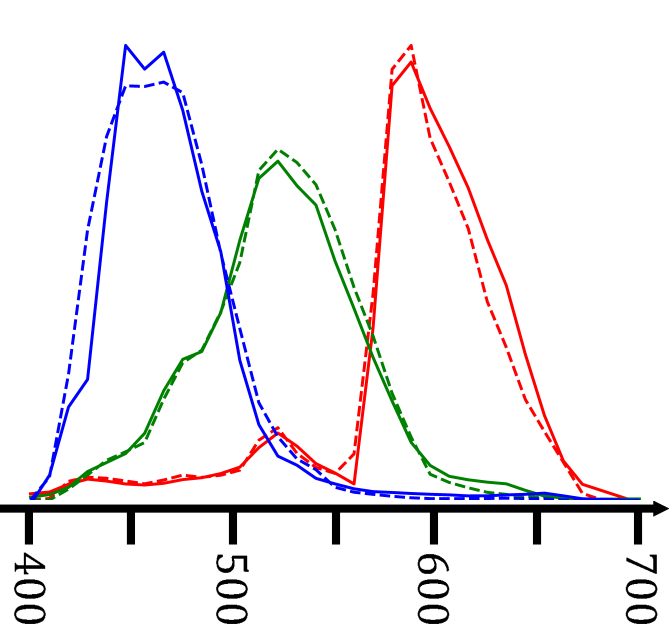}
&
\includegraphics[width=0.333333\linewidth]{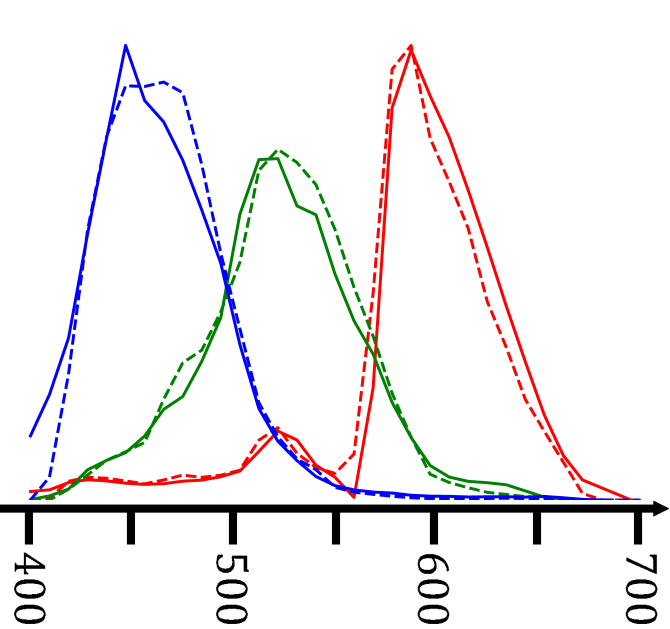}\\
{$\num{5.45E-02}$} & {$\num{5.44E-02}$} & {$\num{5.97E-02}$} \\
\end{tabular}

        \caption{Estimated results of our method using different diffraction grating sheets for \sony with LED~1}
        \label{supfig:real-gratings-comparison}
        \end{figure}

\Fref{supfig:real-leds-comparison} shows the results for \sony using LED~1, LED~2, and LED~3 with Grating~1. All results demonstrate nearly identical camera spectral sensitivities, indicating the stability of our method across different light sources.
\Fref{supfig:real-gratings-comparison} compares the use of different diffraction grating sheets on \sony using LED~1. Overall, the estimated results remain consistent regardless of the diffraction grating choice.

\section{Experiment results of pixel-to-wavelength mapping}
\begin{figure}
    \centering
    \includegraphics[width=\linewidth]{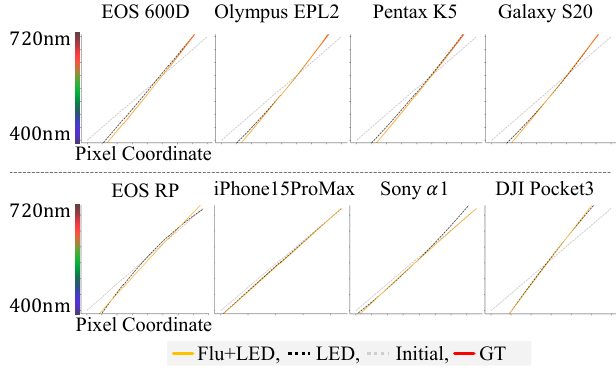}
    \caption{
    Plots of pixel-to-wavelength mapping from the synthetic experiment in the main paper (top row) and the real-world experiment (second row.) 
    The legend is shown in the figure. We put the ground-truth plot only for the synthetic experiment, since we do not have the ground-truth mapping in the real-world setting. The Initial corresponds to the initial guess of the pixel-to-wavelength mapping for Ours~(LED), where $a=0$, $b=f/n$, $c=0$, as described in the main paper.
    $x$-axis corresponds to the pixel coordinate, and $y$-axis to wavelength. Estimation using a non-spiky spectrum can achieve comparable results to estimation using a spiky spectrum. 
    }
    \label{supfig:pix2wave}
\end{figure}

To evaluate mapping estimation using a light source with spiky and non-spiky spectrum, we show the qualitative results of the pixel-to-wavelength mapping by our methods in \fref{supfig:pix2wave}, with quantitative results shown in \Tref{table:pix2wave}. We used the following relative error metric, similar to the main paper:
\begin{equation}
\begin{aligned}
    \mathrm{RE} &= \frac{1}{{\lambda_M - \lambda_m}}{\sqrt{\frac{\norm{\boldsymbol{\lambda}^* - \boldsymbol{\lambda}}_2^2}{N}}}, \\
\end{aligned}
\end{equation}
where \mbox{$\boldsymbol{\lambda}=[g(p_1),\dots, g(p_n)]\in\doubleR^n$} is the corresponding wavelength interpolated by pixel-to-wavelength mapping function $g(p)$, and \mbox{$\boldsymbol{\lambda}$} and      \mbox{$\boldsymbol{\lambda}^*$} is the interpolated wavelength by ground-truth and estimated pixel-to-wavelength mapping function, respectively.

The experiment shows that the pixel-to-wavelength mapping using fluorescent light can achieve around $0.3\%$ error regardless of the target camera sensitivity. While the mapping estimation using an LED has a larger error compared to the fluorescent one, and the error is more sensitive to variations in camera spectral sensitivity due to the relatively low-frequency light spectrum, the results still show at most $2\%$ error. This yields comparable results to the estimation using fluorescent light, even with a rough initial guess, in both the synthetic and real-world experiments.
\begin{table}[t]
\small
    \centering
    \begin{tabular}{@{}c@{}||@{}c@{}c@{}c@{}c@{}}
\makebox[0.200000\linewidth]{} 
 &
 \multirow{2}{*}{\makebox[0.200000\linewidth]{\syneos}}
&
\makebox[0.200000\linewidth]{
Olympus}
&\makebox[0.200000\linewidth]{
Pentax K5}
&
\makebox[0.200000\linewidth]{Sumsung}\\
&
&
\makebox[0.200000\linewidth]{EPL2} & 
& 
\makebox[0.200000\linewidth]{Galaxy S20}\\
\hline
Flu+LED & \num{0.31} & \num{0.31} &  \num{0.31} & \num{0.31}  \\
LED         & \num{2.18} & \num{2.13} &  \num{2.34} & \num{2.10}                          
\end{tabular}
        \caption{Synthetic quantitative results of pixel-to-wavelength mapping. "Fluorescent" corresponds to the estimation using a light source with a spiky spectrum, and "LED" to the estimation with a non-spiky spectrum. The values are scaled by $\num{E2}$ as percentage for display purposes.
        }
    \label{table:pix2wave}
\end{table}

\section{Pixel-to-wavelength mapping representation}
In this section, we provide a rationale for approximating the pixel-to-wavelength mapping $\lambda(p)$ using a quadratic function. 

\begin{figure}[t]
    \centering
    \includegraphics[width=\linewidth]{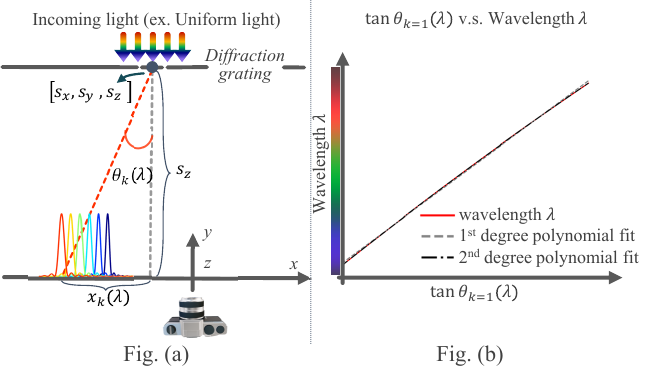}
    \caption{
       (a) Visual illustration of the geometric relation of the diffraction grating and the diffraction angle. (b) The plot of $\lambda$ with respect to $\tan(\theta_{k=1})$. In the visible wavelength domain, \first~or$2^\mathrm{nd}$  degree of polynomial fit can sufficiently represent the original function.
    }
    \label{supfig:grating-eq}
\end{figure}

Assuming an ideal diffraction grating, there is a linear relation between a certain wavelength $\lambda$ and the sin value of \ithEq{k} diffraction angle $\theta_k(\lambda)$ as
\begin{equation}
    \label{eq:maxima_point}
    d \sin{\theta_k(\lambda)} = k\lambda,
\end{equation}
where $d$ is the slit width of the diffraction grating,
and the diffraction angle $\theta_k(\lambda)$ describes the \ithEq{k} intensity maxima position $x_k(\lambda)$ as illustrated in \fref{supfig:grating-eq}~(a).

With $\SI{500}{slits/\milli\meter}$ diffraction grating, which is used in our real-world experiments, for example, the slit width $d$ is $d=\frac{1}{500}\si{\milli\meter}=\SI{2.0E-6}{\meter}$.
Given the visible wavelength domain, which is approximately in $[\SI{380}{\nano\meter}, \SI{720}{\nano\meter}]$, we can compute the possible range of sine of the \first diffraction grating angle as:
\begin{equation}
   \begin{gathered}
   \frac{\num{3.8E-7}}{\num{2E-6}} \le\sin{\theta_{k=1}}\le \frac{\num{7.2E-7}}{\num{2E-6}} \\
   \therefore 0.190 \le\sin{\theta_{k=1}}\le 0.360 \\
   \therefore 0.191 \le \theta_{k=1}(\lambda) \le 0.369.
   \end{gathered}
\end{equation}

\Fref{supfig:grating-eq}~(b) shows the plot of $y=\lambda$ with respect to $\tan{\theta_k(\lambda)}$ where \hbox{$\theta_{k=1}(\lambda) \in [0.191, 0.369]$}, accompanied with the \first~and $2^\mathrm{nd}$ degree of polynomial fit results of  $y=\lambda$. 
It is obvious that the wavelength $\lambda$ and $\tan{\theta_k(\lambda)}$ is almost in the affine relation in the given domain, and at most $2^\mathrm{nd}$ degree of polynomial approximation is good enough to represent the wavelength $\lambda$ using a tangent value of diffraction angle, $\tan{\theta_k(\lambda)}$.

At the same time, $\tan{\theta_k(\lambda)}$ can be described as 
\begin{equation}
\tan{\theta_k(\lambda)} = \frac{x_k(\lambda)}{s_z},
\end{equation}
where
$
 \begin{bmatrix}
    s_x, & s_y, & s_z
\end{bmatrix}\transp\in\doubleR^{3}
$ is the center coordinate of the diffraction grating in the camera coordinate system.
Thus similarly, $\lambda$ can be represented using $2^\mathrm{nd}$ degree of polynomial of $x_k(\lambda).$

Furthermore, assuming the image plane and the diffraction grating sheet are parallel with each other, then we can define the intensity maxima position in the camera coordinate system $\V{x}_k(\lambda)\in\doubleR^3$ as  

\begin{equation}
    \V{x}_k(\lambda) = \begin{bmatrix}
    x_k(\lambda) + s_x\\
    s_y \\
    s_z,
\end{bmatrix}.
\end{equation}
Assuming a pin-hole camera model, the corresponding pixel coordinate along $x$-axis, or along the diffraction direction, $p_k(\lambda)$ can be represented in an affine relation as 
\begin{equation}
    p_k(\lambda) = f_x  \frac{x_k(\lambda)+s_x}{s_z}  + c_x,
\end{equation}
where $f_x$ and $c_x\in\doubleRp$ represent focal length and the image center in the horizontal direction, respectively.

Overall, since $\lambda$ can well represented by $2^\mathrm{nd}$ degree of polynomial of $\tan{\theta_k(\lambda)}$ and since $\tan{\theta_k(\lambda)}$ and the corresponding pixel position $p_k(\lambda)$ is in an affine relation to $\tan{\theta_k(\lambda)}$, wavelength $\lambda$ can be represented 
by quadratic function of the pixel coordinate $p(\lambda)$, and pixel-to-wavelength mapping $\lambda = g(p)$ is represented as Eq.~(13) in the main paper.

        \begin{figure*}[t]
        \small
        \centering
        \begin{tabular}{@{}c@{\hspace{0.1em}}c@{\hspace{0.1em}}c@{\hspace{0.1em}}c@{\hspace{0.1em}}c@{}}
\makebox[0.19\linewidth]{1 Patch}
&
\makebox[0.19\linewidth]{5 Patches}
&
\makebox[0.19\linewidth]{10 Patches} 
& 
\makebox[0.19\linewidth]{25 Patches} 
& 
\makebox[0.19\linewidth]{N-band filter} \\
\includegraphics[width=0.19\linewidth, height=0.147\linewidth]{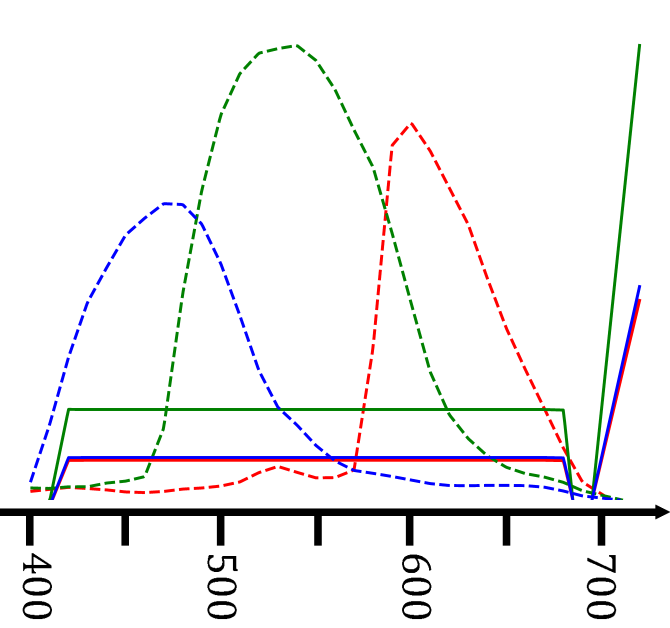}
&
\includegraphics[width=0.19\linewidth, height=0.147\linewidth]{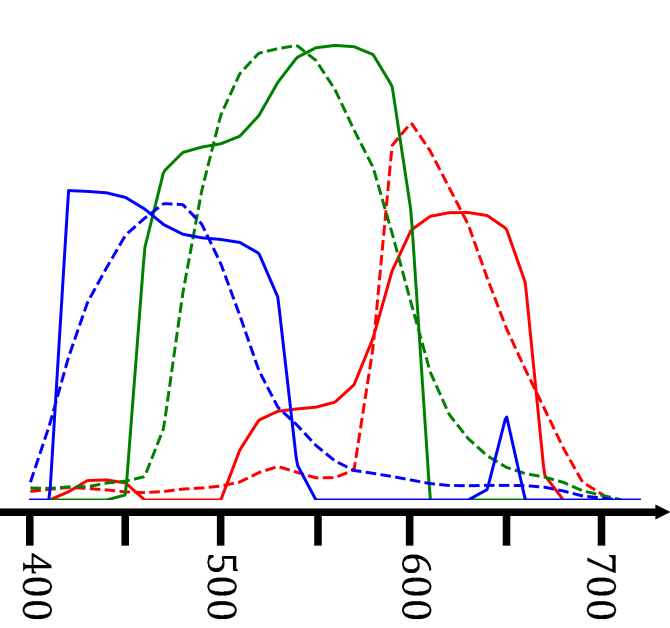}
&
\includegraphics[width=0.19\linewidth, height=0.147\linewidth]{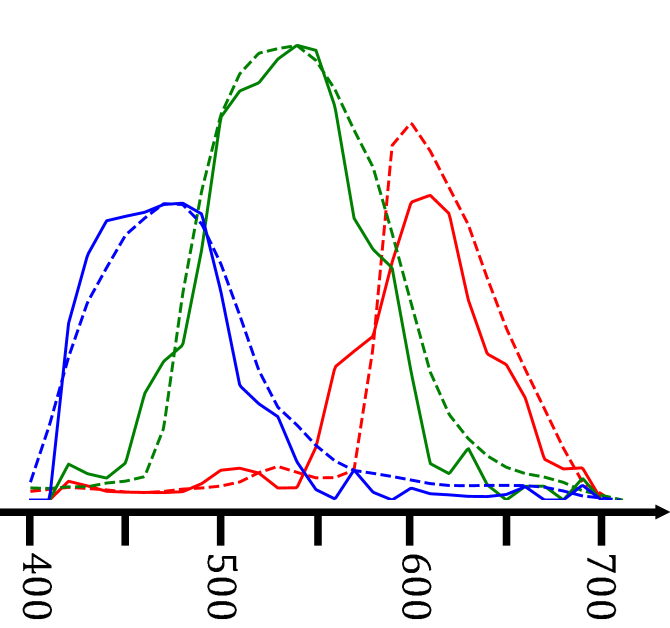}
&
\includegraphics[width=0.19\linewidth, height=0.147\linewidth]{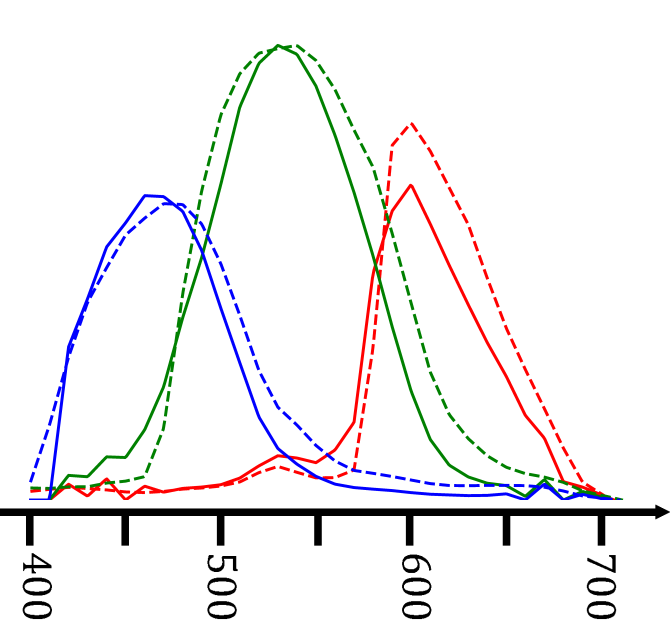}
&
\includegraphics[width=0.19\linewidth, height=0.147\linewidth]{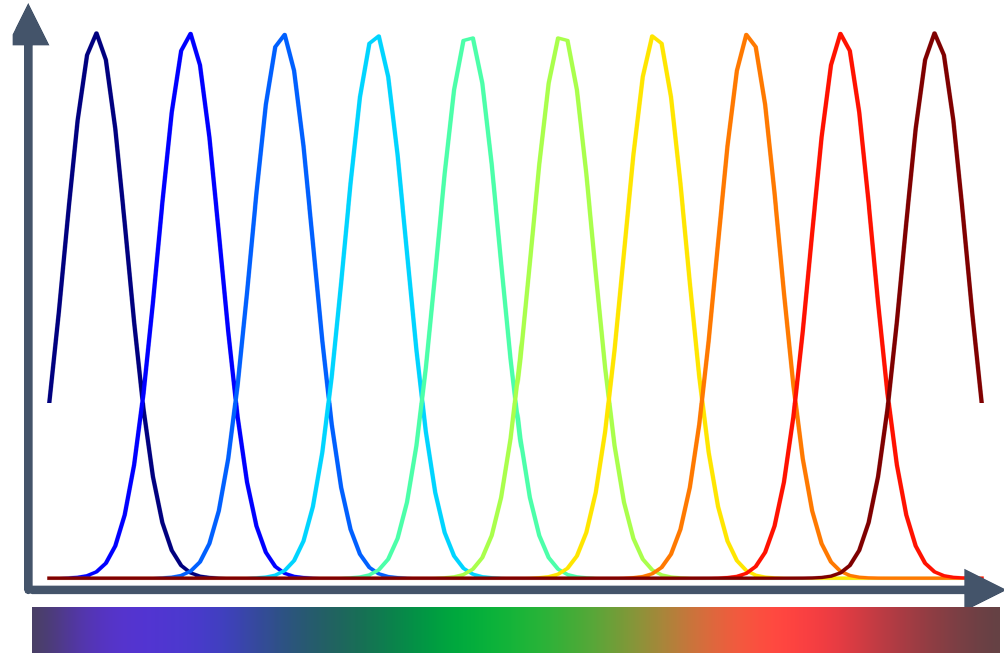}\\
{$\num{4.43E-01}$} & {$\num{1.47E-01}$} & {$\num{9.69E-02}$} & $\num{8.20E-02}$ \\
        \end{tabular}
        \caption{Estimation results of the existing diffraction grating-based method~\cite{toivonen2020practical} using different numbers of patches in the transmissive color chart. 
        Since there is no known dataset for transmissive filters, we synthesized $N$-band transmissive filters. We implemented them using shifted Gaussian distributions, ensuring that the visible wavelength range was always fully covered by the $N$ filters. The rightmost figure shows an example filter spectrum for the $N=10$ case.
}
        \label{supfig:existing_comparison}
        \end{figure*}

\section{Basis representations for grating efficiency}
We define a Fourier basis for the grating efficiency function $\V{B}_\eta = [\V{b}_k]_{k=1,...,t} \in \doubleR^{f\times t}$ as follows:
\begin{equation}
    \V{b}_k = \left\{
    \begin{array}{ll}
    \cos\left(2\pi k f_i \right) & \text{if } k \mbox{ is odd}, \\
    \sin\left(2\pi k f_i \right) & \text{if } k \mbox{ is even and } k\neq 0, \\
    1 & \text{if } k = 0, \\
    \end{array}
    \right.
\end{equation}
where $t$ denotes a number of basis, $\lambda_l, \lambda_u$ are the lower and upper bound of the wavelength range, respectively, and {$f_i = \frac{\lambda_i}{\lambda_u - \lambda_l}$}.

\section{Estimation details}
Here we detail the vectors and matrices that appear in Eq.~(12) of the main paper:
\begin{equation}
    \begin{gathered}
    \V{x}^* = \argmin_{\V{x}} \left\| \V{A}_\mathrm{dif} 
\V{x} \right\|_2^2~~
\mathrm{s.t.}~~ 
    \begin{bmatrix}
        \V{0} & \V{A}_\mathrm{dir}
    \end{bmatrix}    
\V{x} = 
\V{m}_\mathrm{dir}.
    \end{gathered}
    \label{seq:problem_def}
\end{equation}
In the constraint, $\V{A}_\mathrm{dir}$ is a $3\times 3b_s$ dimensional matrix written as
\begin{equation*}
\small
        \V{A}_\mathrm{dir} = \begin{bmatrix}
        \V{e}^\top \V{B}_{s(R)} & 0 & 0 \\
        0 & \V{e}^\top \V{B}_{s(G)} & 0 \\
         0 & 0 & \V{e}^\top \V{B}_{s(B)} \\
        \end{bmatrix},
\end{equation*}
where $\V{B}_{s(R)}$, $\V{B}_{s(G)}$, and $\V{B}_{s(B)}$ are basis matrices for each RGB color channel.
The accompanying zero matrix \V{0} is $3 \times b_\eta$ dimensional.

In the objective, $\V{A}_\mathrm{dif}$ becomes a $3f\times (b_\eta + 3b_s)$ dimensional matrix as
\begin{equation*}
        \V{A}_\mathrm{dif} = 
\begin{bmatrix}
\diag{\V{a}_{(R)}}\V{B}_\eta & -\V{B}_{s(R)} & 0 & 0 \\
\diag{\V{a}_{(G)}}\V{B}_\eta & 0 & -\V{B}_{s(G)} & 0 \\
\diag{\V{a}_{(B)}}\V{B}_\eta & 0 & 0 & -\V{B}_{s(B)}
\end{bmatrix},
\end{equation*}
where vector $\V{a}_{(c)}$ is defined as 
\begin{equation*}
\V{a}_{(c)} = \diag{\V{e}^{-1}} \V{W}^\dagger \V{m}_{\mathrm{dif}{(c)}}
\end{equation*}
for color channel $c \in \{R,G,B\}$, and $\V{m}_{\mathrm{dif}{(c)}}$ is a vector of diffracted light observations in color channel $c$.
 \mbox{$\V{x}=\begin{bmatrix} \V{c}_\eta\transp, & \V{c}_{s(R)}\transp, & \V{c}_{s(G)}\transp, & \V{c}_{s(B)}\transp \end{bmatrix}\transp\in\doubleR^{b_\eta + 3b_s}$} is the vector of unknowns in total.

To solve this problem, we introduce Lagrange multiplier $\boldsymbol{\mu}\in\doubleR^{3}$ whose dimension corresponds to the number of equations in the linear constraints from the direct light observation $\V{A}_\mathrm{dir}$. With the Lagrange multiplier $\boldsymbol{\mu}$, the Lagrangian $\mathcal{L}$ is defined as: 
\begin{equation*}
    \mathcal{L} = \norm{\V{A}_\mathrm{dif}{\V{x}}}_2^2 + \boldsymbol{\mu}\transp\left(\begin{bmatrix}\V{0} &\V{A}_\mathrm{dir}\end{bmatrix}{\V{x}} - \V{m}_\mathrm{dir}\right),
\end{equation*}
whose stationary point should satisfy $\frac{\partial \mathcal{L}}{\partial \V{x}}=\V{0}$ and $\frac{\partial \mathcal{L}}{\partial\boldsymbol{\mu}} = \V{0}$.
It results in the following linear system of equations: 
\begin{equation*}
    \V{A}
    \begin{bmatrix}
        \V{x}  \\
        \boldsymbol{\lambda} 
    \end{bmatrix}  
    = \V{b},
\end{equation*}
where 
\begin{equation*}
\small
    \begin{aligned}
        \V{A} &= \begin{bmatrix}
2\V{A}_\mathrm{dif}\transp\V{A}_\mathrm{dif} & \begin{bmatrix}\V{0} &\V{A}_\mathrm{dir}\end{bmatrix}\transp\\
        \begin{bmatrix}\V{0} &\V{A}_\mathrm{dir}\end{bmatrix} & \V{0} 
    \end{bmatrix}\in\doubleR^{(b_\eta + 3b_s + 3)\times (b_\eta + 3b_s + 3)}  \\
    \V{b} &= 
    \begin{bmatrix}
        \V{0}  \\
        \V{m}_\mathrm{dir}
    \end{bmatrix}\in\doubleR^{b_\eta + 3b_s + 3}
    \end{aligned}.
\end{equation*}
The estimates of $\V{x}^*$ and $\boldsymbol{\mu}^*$ is found by $\V{A}^{-1}\V{b}$ if $\V{A}^{-1}$ exists. Since the system in \eref{seq:problem_def} is already overdetermined and sufficiently robust, we found that the explicit positivity constraints are unnecessary, as confirmed by our experiments.
The full estimation of pixel-to-wavelength mapping and camera spectral sensitivity takes about two to four minutes per camera on a single core of an Intel Xeon CPU.

\section{Existing diffraction grating-based method}
The previous method~\cite{toivonen2020practical} uses diffracted observations from multiple light sources and images of a transmissive color chart. In theory, their method can be applied in combination with our estimation method of the pixel-to-wavelength mapping. Specifically, we can use a diffracted observation from a single light source and treat the observation of the direct light as a single patch of the transmissive color chart.
However, based on our preliminary experiments, we found that their solution method based on stochastic gradient descent using the ADAM optimizer is unstable. 

\Fref{supfig:existing_comparison} shows their estimation results by varying the number of patches in the transmissive color chart. As stated above, the proposed setup corresponds to the single-patch case, and we confirm that stable estimation with~\cite{toivonen2020practical} requires using more than $10$ patches.

As in the one-patch result from the above preliminary experiment, the estimation in the real-world experiments shows $\mathrm{RE}=0.58\mathrm{{~to~}}0.65$ when evaluated on the same input as ours.
This emphasizes the effectiveness of the proposed solution method using a linear solution method, which works well for the single-patch case.

\section{Slit width impact}
This section outlines the conditions under which one-to-one pixel-to-wavelength mapping can be assumed.
One-to-one pixel-to-wavelength mapping is ensured by two conditions: 
a)~diffraction resolvance $\Delta\lambda\!=\!\lambda_f / N$ (with $N$ being the number of slits) determined by the Rayleigh criterion must be finer than the ground-truth resolvance $\Delta\lambda_\mathrm{GT}$, and
b) the diffracted pattern must span at least $f$ pixels, \ie $n \ge f$.
Given $\Delta\lambda_\mathrm{GT} \approx \SI{10}{\nano\meter}$ and $\lambda_f = \SI{720}{\nano\meter}$ from CamSPECS~XL, condition (a) holds when $N \ge 100$.
Our gratings ($500$ or $1000$~\si{slits\per\milli\meter}) meet this, and we align the camera so that the pattern spans at least $f = 31$ pixels. 

\begin{figure}[t]
    \centering
        \includegraphics[width=\linewidth]{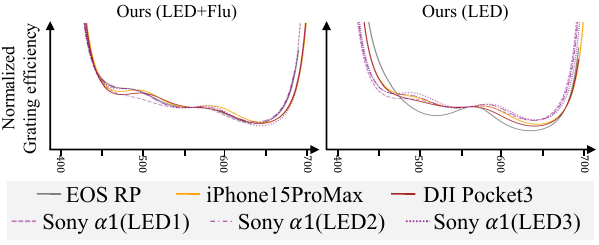}
        \caption{Normalized grating efficiency plot from Ours~(LED+Flu) and Ours~(LED).} 
        \label{sfig:grating_and_comparion}
\end{figure}
\begin{figure}[t]
    \centering
        \includegraphics[width=\linewidth]{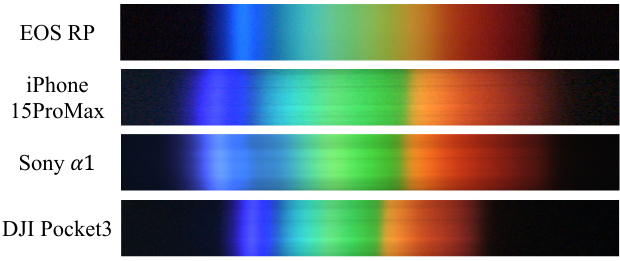}
        \caption{Diffraction observations in the real-world experiments.} 
        \label{sfig:diffraction_obs}
\end{figure}

\section{Consistency of efficiency estimation}
To assess the consistency of the diffraction grating efficiency estimation across different scenes, we show a comparison of the estimated efficiencies.
\Fref{sfig:grating_and_comparion} shows the estimated efficiencies for the scenes in the real-world experiments~(see Fig.7 in the main paper), which employ different cameras 
with varying relative positions between the camera and the slit, resulting in varying pixel-to-wavelength mappings.
The mean cosine similarities of all estimations are $0.97$ (LED+Flu) and $0.96$ (LED), indicating the consistency of the proposed efficiency estimation method. 

\section{Diffracted light observation in the real-world experiments}
\Fref{sfig:diffraction_obs} shows the input diffracted light observations in the experiments. 
The observed diffraction patterns vary with each camera’s spectral sensitivity, enabling us to estimate those sensitivities.
{
    \small
    \bibliographystyle{ieeenat_fullname}
    \bibliography{ref}
}